# Self-service Data Classification Using Interactive Visualization and Interpretable Machine Learning

Sridevi Narayana Wagle, Boris Kovalerchuk

Department of Computer Science, Central Washington University, USA

**Abstract:** Machine learning algorithms often produce models considered as complex black-box models by both end users and developers. Such algorithms fail to explain the model in terms of the domain they are designed for. The proposed *Iterative Visual Logical Classifier* (IVLC) is an interpretable machine learning algorithm that allows end users to design a model and classify data with more confidence and without having to compromise on the accuracy. Such technique is especially helpful when dealing with sensitive and crucial data like cancer data in the medical domain with high cost of errors.

With the help of the proposed interactive and lossless multidimensional visualization, end users can identify the pattern in the data based on which they can make explainable decisions. Such options would not be possible in black box machine learning methodologies. The interpretable IVLC algorithm is supported by the *Interactive Shifted Paired Coordinates Software System* (SPCVis). It is a lossless multidimensional data visualization system with user interactive features. The interactive approach provides flexibility to the end user to perform data classification as self-service without having to rely on a machine learning expert. Interactive pattern discovery becomes challenging while dealing with large data sets with hundreds of dimensions/features. To overcome this problem, this chapter proposes an automated classification approach combined with new *Coordinate Order Optimizer* (COO) algorithm and a *Genetic algorithm*. The COO algorithm automatically generates the coordinate pair sequences that best represent the data separation and the genetic algorithm helps optimizing the proposed IVLC algorithm by automatically generating the areas for data classification. The feasibility of the approach is shown by experiments on benchmark datasets covering both interactive and automated processes used for data classification.

## 1. Introduction

The exponential growth in the Machine Learning has resulted in smart applications making critical decisions without human intervention. However, people with no technical background find it often difficult to rely on machine learning models, since many of them are black-boxed [Feurer et al, 2019].

Interpretability plays a crucial role in deciphering behind-the-scenes actions in a machine learning algorithm. These interpretable machine learning techniques when combined with data visualization and analytics unfold numerous ways to discover deep hidden patterns in the data [Kovalerchuk, 2018]. Interpretable machine learning models are transparent clearly explaining the technique behind the model prediction. This gives more clarity to the end user who can rely on the model with more confidence compared to black-box machine learning models [Hutter et al, 2019].

When it comes to data visualization, most of the visualization techniques used to display multidimensional data are lossy and irreversible [Kovalerchuk, 2018]. With the help of new lossless data visualization, it is now possible to maintain structural integrity of the data. Moreover, these lossless techniques are completely reversible [Kovalerchuk, 2018, Kovalerchuk, Ahmad, Teredesai, 2021].

In this chapter, we focus on interpretable data classification techniques using a combination of interactive data visualization and analytical rules. The Shifted Paired Coordinates (SPC) system [Kovalerchuk, 2018] is a lossless and a more compact way to visualize multidimensional data when compared to other lossless data visualization like Parallel Coordinates. SPC allows discovering patterns more efficiently since the number of lines required to display the data is reduced by half [Kovalerchuk, 2018].

However, as the size of the data increases, pattern discovery becomes challenging due to occlusion. In order to reduce occlusion and expose hidden patterns, data representation needs to be reorganized. This can be achieved by applying interactive techniques such as changing the order of coordinates, swapping within the coordinate pair, etc. These interactive capabilities allow the end-users to intervene and optimize the classification model generated by the machine thereby improving the overall model performance. To further leverage the classification technique, areas within the coordinate pairs are discovered either interactively or automatically, unravelling the deep hidden patterns in the data. Analytical rules are then built on the areas discovered to classify the data.

The data classification approach using analytical rules is implemented using IVLC algorithm [Wagle, Kovalerchuk, 2020] wherein the areas and analytical rules are generated in iterations until all the data are covered. This implementation works for smaller datasets, the areas and analytical rules can be generated interactively by the end users. This chapter expands this work to large datasets by adding automation and more visual operations.

For the classification of larger data with hundreds of dimensions, interactive methods alone do not suffice. Due to high occlusion for such datasets, discovery of patterns purely on interactive methods becomes challenging where deep hidden patterns cannot be detected by humans. In this chapter, automation for the classification approach is implemented in two stages:

*First Stage*: Order of coordinates are optimized using COO algorithm. It is used to find the best coordinate orders for the Shifted Paired Coordinates System where the data separation can be visually identified along the vertical axis of each coordinate pair.

*Second Stage*: This stage involves generating the areas within the coordinate pairs with high purity or fitness i.e., areas with high density of data belonging to same class are generated. The genetic algorithm is used for the automatic generation of areas with high purity. It involves generation of random areas and that are mutated and altered over generations to obtain the areas with high purity or fitness [Banzhaf et al, 1998].

Several experiments are conducted on benchmark datasets like Wisconsin Breast Cancer (WBC), Iris, Seeds and Air Pressure System (APS) in Scania Trucks. The experiments are performed using both interactive and automated approach using 10-fold cross validation with worst case heuristics [Kovalerchuk, 2020] where the initial validation set contains the data from one class overlapping with another class and vice versa. The results obtained with both interactive and automated techniques were on par with the published results in other studies.

This chapter offers a detailed explanation of the following:

- Interactive Shifted Paired Coordinates Visualization System (SPCVis);
- Different interactive techniques used in SPCVis;
- Iterative Visual Logical Classifier (IVLC) algorithm [Wagle, Kovalerchuk, 2020];
- Automation of classification using Coordinate Order Optimizer and Genetic Algorithms;
- Worst case heuristic approach for 10-fold cross validation;
- Experiments with benchmark datasets;
- Comparing the results with published results in other studies on benchmark datasets.

## 2. Interactive Shifted Paired Coordinates System

The Shifted Paired Coordinates (SPC) [Kovalerchuk B, 2018] visualize data losslessly using coordinate pairs where each pair of data is represented on a two-dimensional plane. Fig. 1 represents an 8-D data (8, 1, 3, 9, 8, 3, 2, 5) using SPC, where pair (8,1) is visualized in $(X_1, X_2)$, the next pair (3,9) is visualized in $(X_2, X_4)$ and so on. Then these points are connected to form a graph.

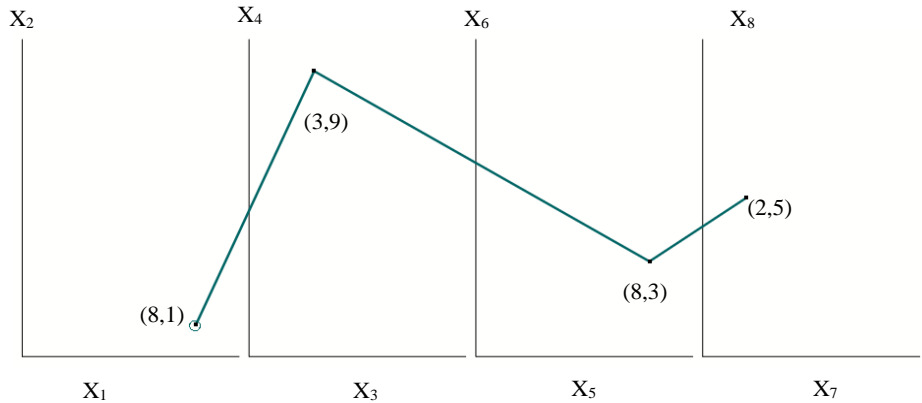

**Fig. 1.** Representation of 8-D data (8, 1, 3, 9, 8, 3, 2, 5) in SPC.

Same data can be displayed in multiple ways using different combinations of pairs of coordinates. Figs. 2a and 2b represent the same data with $(X_8, X_1)$, $(X_3, X_7)$, $(X_5, X_2)$ and $(X_6, X_4)$ sequence of coordinates and $(X_2, X_6)$, $(X_3, X_7)$, $(X_8, X_1)$ and $(X_5, X_4)$ sequence of coordinates respectively.

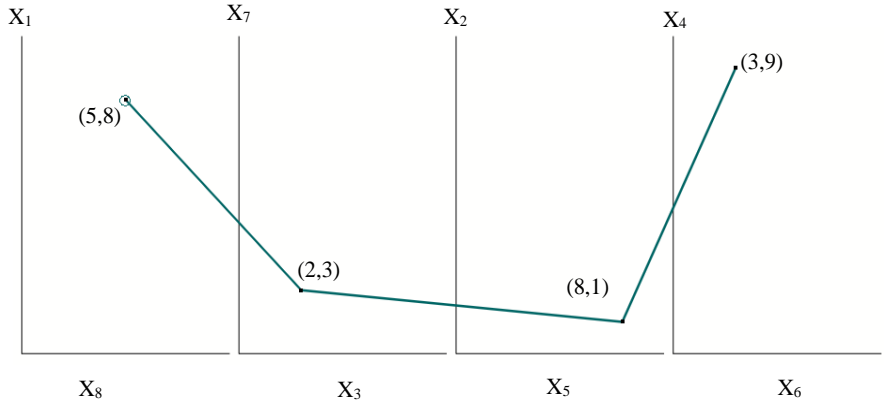

**(a).** 8-D data with $(X_8, X_1)$, $(X_3, X_7)$, $(X_5, X_2)$ and $(X_6, X_4)$ coordinate pair sequence.

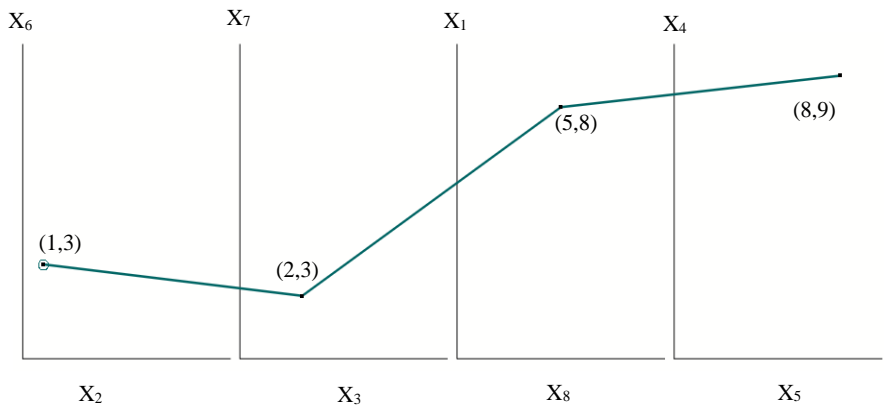

**(b).** 8-D data with $(X_2, X_6)$, $(X_3, X_7)$, $(X_8, X_1)$ and $(X_5, X_4)$ coordinate pair sequence.

**Fig. 2.** Representation of 8-D data $(8, 1, 3, 9, 8, 3, 2, 5)$ in SPC with different coordinate pair sequences.

Fig. 3 represents a real-world dataset from UCI Machine Learning Repository [Dua et al, 2019] consisting of 683 cases of breast cancer data. Coordinate $X_2$ is duplicated to get even number of coordinates. Although Shifted Paired Coordinates system provides lossless data visualization, discovering patterns in the data becomes challenging due to occlusion. To overcome this challenge, IVLC algorithm is used comprising of interactive controls to reorient the data to make the pattern discovery easier along with analytical rules to classify the data.

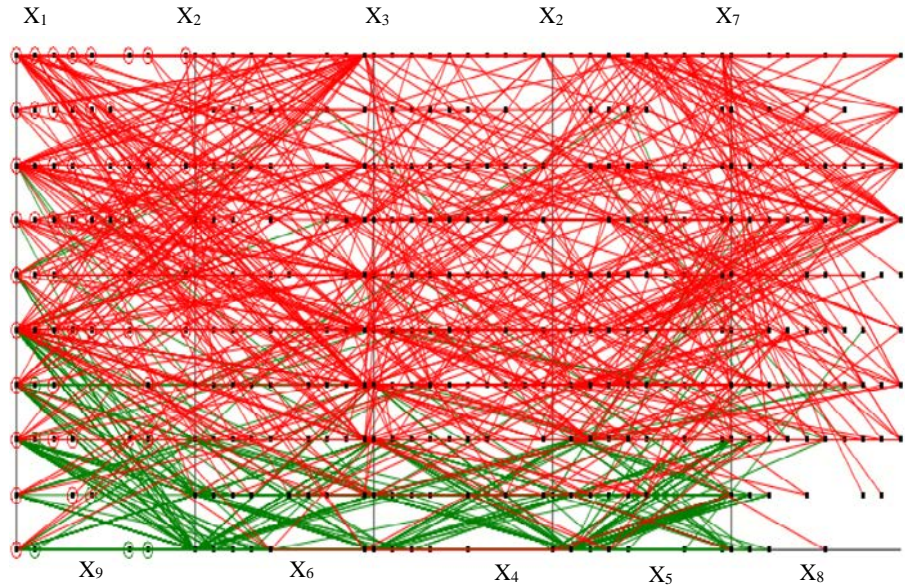

(a). WBC data with red class on top.

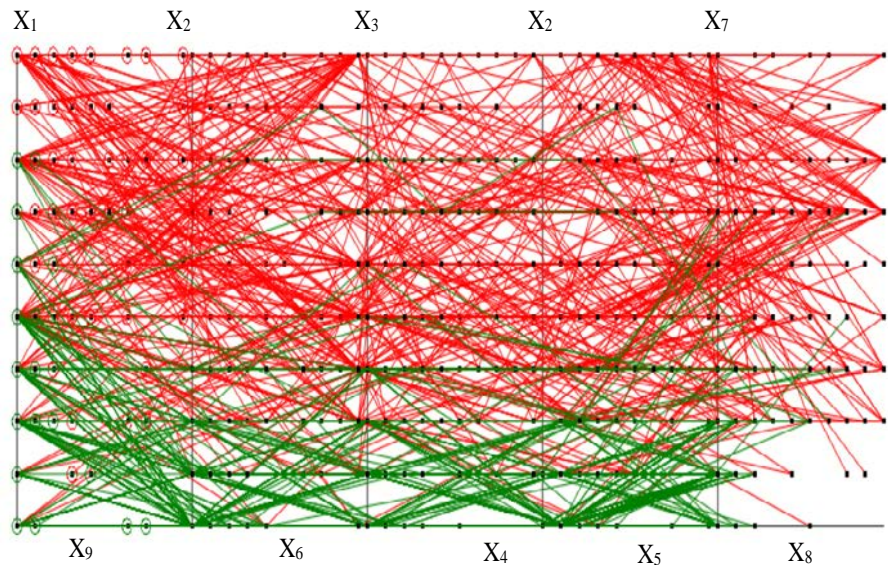

(b). WBC data with green class on top.

**Fig. 3.** Wisconsin Breast Cancer (WBC) 9-D dataset visualized in SPCVis.

The lossless n-D visualization is achieved by representing data in *Interactive Shifted Paired Coordinates System* (SPCVis). Reordering the coordinates is one of the interactive features provided by the SPCVis software system. Using this feature, the coordinates are reordered in such a way that the class separation is prominent along the vertical coordinates. The discovering of coordinates to get good separation of classes is performed interactively by the user.

There are several interactive features provided to the end user like reversing data, non-linear scaling etc. For instance, if **x** is an n-D point where **x** = $(x_1, x_2, x_3, \ldots\ldots x_n)$, the reverse of $x_1$ would display the data as $(1 - x_1, x_2, x_3, \ldots\ldots x_n)$. Also, SPCVis software system provides the user ability to click and drag the whole $(X_i, X_j)$ plot to desired location until occlusion is reduced. Another interactive control allows a user to display data of user selected class on top of another class. Fig. 3a displays WBC data with red class on top and Fig. 3b with green class on top. This helps the user to observe the pattern of individual classes more clearly.

Non-linear scaling is an interactive feature provided by the SPCVis Software where only a part of the user selected coordinate is scaled differently. The generalized formula for an n-D point $x_j$ is given in equation (1),

$$x_j' = \begin{cases} x_j\,, & \text{if } x_j < k \\ x_j + r \times graphWidth, & \text{if } k \leq x_j < 1 \end{cases} \tag{1}$$

here $k$ is a constant and $0 < k < 1$, $r$ is the resolution of the data i.e., the shortest distance between the data points. The value of $k$ is set by the user. The data used for SPC are normalized to [0, 1].

Fig. 4 displays the breast cancer data set after applying the non-linear scaling at $r = 0.1$, $k = 0.6$ on $X_1$ and $k = 0.3$ on $X_2$, $X_5$ and $X_7$ coordinates.

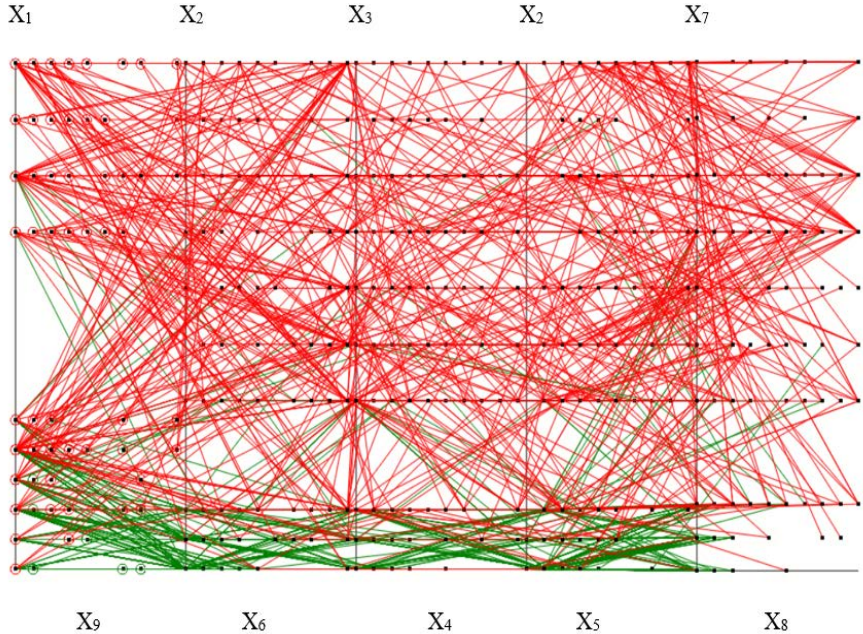

**Fig. 4.** WBC 9-D Data after Non-Linear Scaling on all the Vertical Coordinates.

Another interactive feature provided by SPCVis software is Non-orthogonal coordinate system that has a coordinate inclined at an angle other than 90° with respect to the other coordinate. Fig. 5 displays a simple 2-D graph with Y coordinate inclined at an angle of 30° with respect to its previous Y coordinate. Fig. 6 displays non-orthogonal coordinate representation with horizontal coordinates $X_8$ and $X_5$ inclined at -30°.

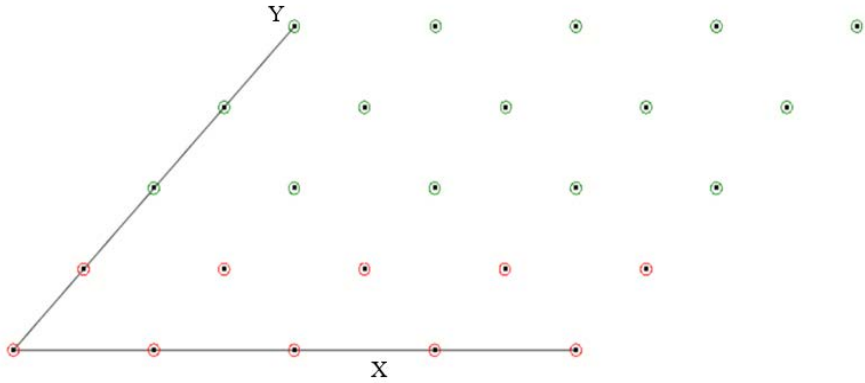

**Fig. 5.** Non-Orthogonal Display of 2-D Data (Y=30°).

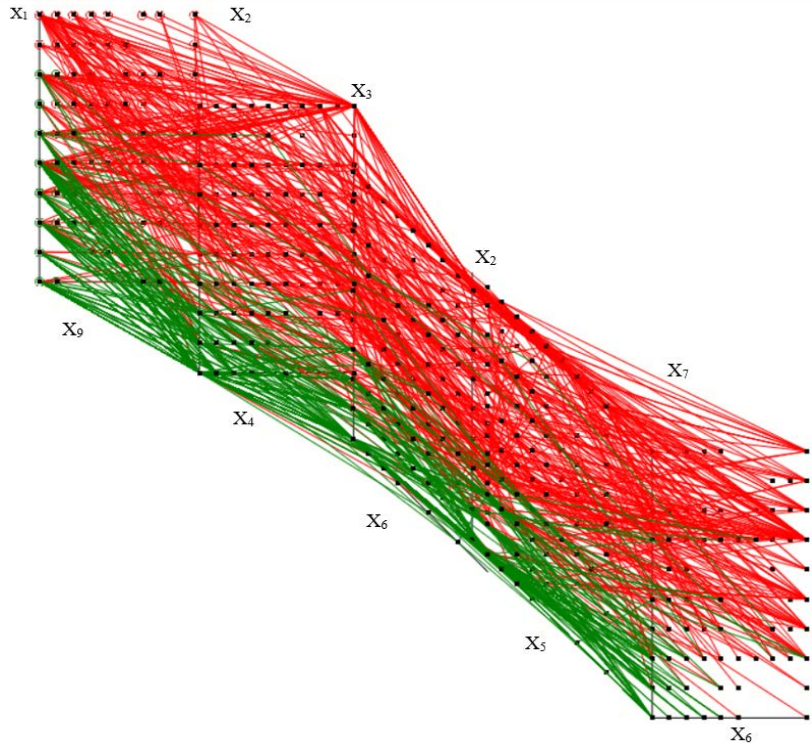

**Fig. 6.** Non-Orthogonal Display of WBC 9-D Data ($X_8$ and $X_5$ inclined at -30°).

Interactive controls like non-linear scaling, non-orthogonal coordinates etc. allow improving visual discrimination of classes. However, interactive visualization alone does not completely perform the data separation. It only provides a base for it.

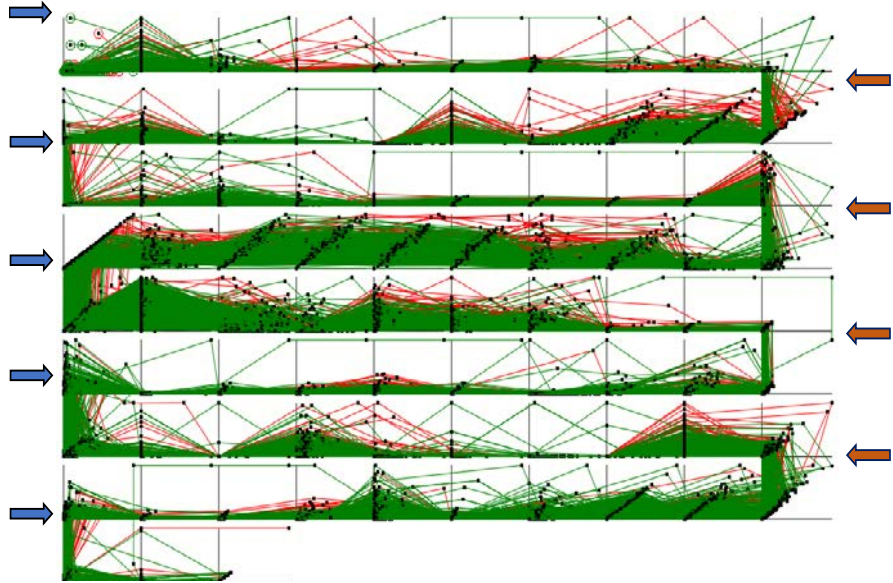

**(a).** APS data with green class on top.

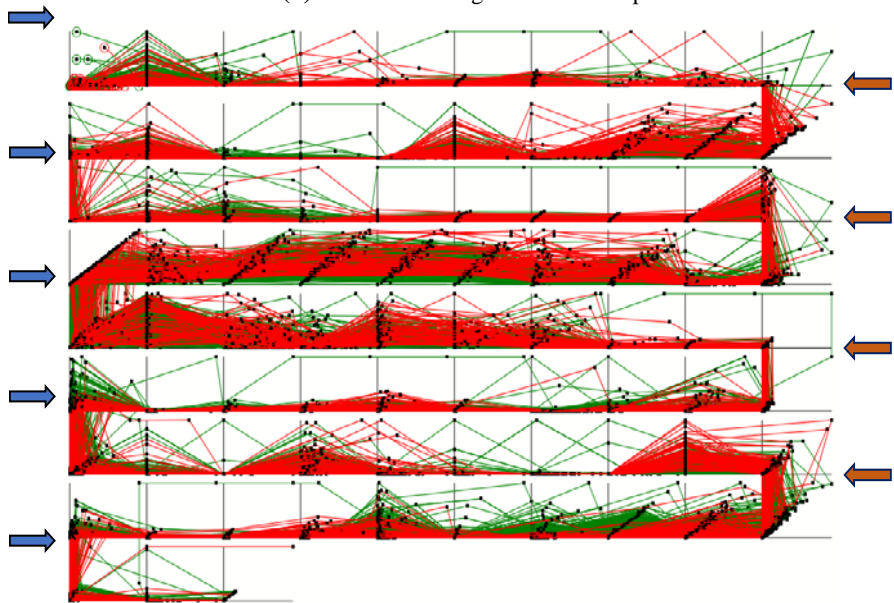

**(b).** APS data with red class on top.

**Fig. 7.** APS Failure at Scania Trucks (166 dimensions) visualized in Serpent Coordinate System.

This chapter uses the IVLC algorithm that generates analytical rules to perform further class separation after reordering the coordinates. This algorithm generates these rules mainly using the threshold values generated from non-linear scaling. The rules belong to the class of rules proposed in [Kovalerchuk, Gharawi, 2018].

Visualizing data with larger dimensions become challenging in SPC. To display data of larger dimensions, a modified version of the SPC called as **Serpent Coordinate System (SCS)** is proposed. It is visualized in a grid like structure to accommodate all the dimensions on a single screen. Air Pressure System (APS) for Scania Trucks [Dua et al, 2019] consists of 2 classes and 170 dimensions wherein 4 dimensions we removed since all the data points under those columns were 0 and were not informative. The data with 166 dimensions are displayed in Fig. 7 and the coordinate labels corresponding each coordinate pair are displayed in Table 1.

Table 1. Coordinate Labels for Serpent Coordinate System (SCS) for Figs. 7a and 7b.

| | | | | | | |
|---|---|---|---|---|---|---|
| $(X_1, X_2)$ | $(X_3, X_4)$ | $(X_5, X_6)$ | .... | $(X_{15}, X_{16})$ | $(X_{17}, X_{18})$ | $(X_{19}, X_{20})$ |
| $(X_{21}, X_{22})$ | $(X_{23}, X_{24})$ | $(X_{25}, X_{26})$ | .... | $(X_{35}, X_{36})$ | $(X_{37}, X_{38})$ | $(X_{39}, X_{40})$ |
| $(X_{41}, X_{42})$ | $(X_{43}, X_{44})$ | $(X_{45}, X_{46})$ | .... | $(X_{55}, X_{56})$ | $(X_{57}, X_{58})$ | $(X_{59}, X_{60})$ |
| $(X_{61}, X_{62})$ | $(X_{63}, X_{64})$ | $(X_{65}, X_{66})$ | .... | $(X_{75}, X_{76})$ | $(X_{77}, X_{78})$ | $(X_{79}, X_{80})$ |
| $(X_{81}, X_{82})$ | $(X_{83}, X_{84})$ | $(X_{85}, X_{86})$ | .... | $(X_{95}, X_{96})$ | $(X_{97}, X_{98})$ | $(X_{99}, X_{100})$ |
| $(X_{101}, X_{102})$ | $(X_{103}, X_{104})$ | $(X_{105}, X_{106})$ | .... | $(X_{115}, X_{116})$ | $(X_{117}, X_{118})$ | $(X_{119}, X_{120})$ |
| $(X_{121}, X_{122})$ | $(X_{123}, X_{124})$ | $(X_{125}, X_{126})$ | .... | $(X_{135}, X_{136})$ | $(X_{137}, X_{138})$ | $(X_{139}, X_{140})$ |
| $(X_{141}, X_{142})$ | $(X_{143}, X_{144})$ | $(X_{145}, X_{146})$ | .... | $(X_{155}, X_{156})$ | $(X_{157}, X_{158})$ | $(X_{149}, X_{160})$ |
| $(X_{161}, X_{162})$ | $(X_{163}, X_{164})$ | $(X_{165}, X_{166})$ | | | | |

.

# 3. Iterative Visual Logical Classifier Algorithm and worst-case *k*-fold Cross Validation

## 3.1. Iterative Visual Logical Classifier Algorithm

Below we present the *Iterative Visual Logical Classifier* (IVLC) algorithm that classifies data in iterations. As discussed in the previous section, once we reorder the coordinates to find good vertical separation and obtain the vertical threshold values from non-linear scaling, the analytical rules are generated interactively based on these threshold values. This process of reordering and generating analytical rules are continued until we cover all the data in given dataset. The algorithm generates a set of interpretable analytical rules for data classification. All steps can be conducted by the end user as a self-service. The steps performed for the classifier are:

**Step 1**: Reorder the coordinates to find a good vertical separation for classes and perform non-linear scaling to get the threshold values along the vertical coordinates. Recording of coordinates and non-linear scaling is performed interactively using SPCVis software system.

**Step 2**: Generate the analytical rules mainly based on the threshold values obtained from non-linear scaling from the previous step. For example, if we denote the set of areas generated as $R_{class1}$ and $R_{class2}$, then the classification rules for n-D point $\mathbf{x} = (x_1, x_2, x_3, \ldots, x_n)$ are:

$$\text{If } x_i \in R_{class1}, \text{ then } \mathbf{x} \in \text{ class 1} \tag{2}$$

$$\text{If } x_j \in R_{class2}, \text{ then } \mathbf{x} \in \text{ class 2} \tag{3}$$

**Step 3**: The data that do not follow the rules generated in step 2 are used as input for next step. Also, in this step, the analytical rules can be tuned to avoid overgeneralization [Kovalerchuk, Grishin, 2018]. Fig. 8b displays the generation of $R_1$ area where a large part of the area is empty. The area can be reduced by generating the area $R_1$ of smaller dimension to avoid overgeneralization (see Fig. 8c).

**Step 4**: Repeat the steps above until all the data are covered.

The output of Iterative Visual Logical Classifier algorithm results in series of rectangular areas. The outputs of IVLC algorithm are displayed in Fig. 8.

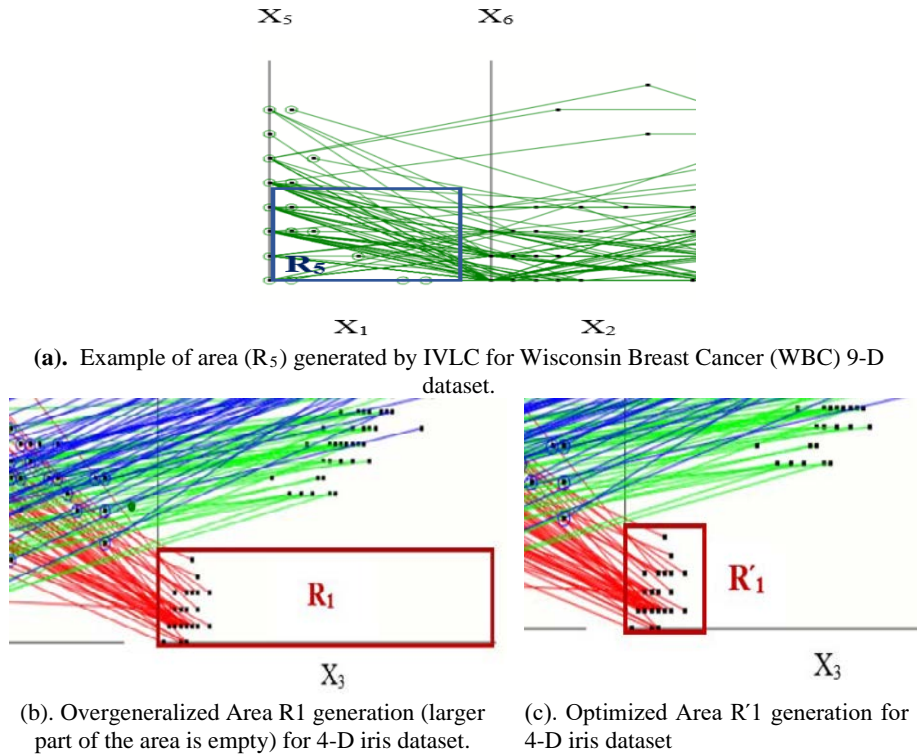

**(a).** Example of area ($R_5$) generated by IVLC for Wisconsin Breast Cancer (WBC) 9-D dataset.

(b). Overgeneralized Area R1 generation (larger part of the area is empty) for 4-D iris dataset.

(c). Optimized Area R'1 generation for 4-D iris dataset

Fig. 8. Outputs of Iterative Visual Logical Classifier (IVLC) Algorithm.

### 3.2. Model Evaluation with Worst-Case *k*-fold Cross Validation Approach

Although Cross Validation is a common technique used for model evaluation, it comes with its own challenges. Due to the random split of training and validation data, we might observe a bias in the estimated average error rate. Also, if we consider all the possible splits, it becomes computationally challenging to find all the combinations since the number of splits grows exponentially with the number of given data points [Kovalerchuk, 2020].

In order to overcome this challenge, we use a worst case heuristics technique to split the data into training and validation sets in *k*-fold cross validation. The worst case fold contains the data of one class that are similar to cases of the opposing class [Kovalerchuk, 2020] making classification more challenging with higher number of misclassification than in the traditional random *k*-fold cross validation. If the algorithm produces high accuracy in the worst-case fold, then the average case accuracy produced by the traditional random *k*-fold cross validation is expected to be greater.

In this chapter, the worst-case fold is extracted using visual representation of the data in Shifted Paired Coordinates System. As already mentioned, the data are displayed in such a way that they tend to be separated along the vertical axes in the SPC. For instance, if the dataset contains class *A* and class *B* with class *B* at the bottom and class *A* at the top in SPC visualization, then the worst-case validation split contains the cases with class *A* displayed on the bottom along with class *B* and vice versa. Since 10-fold cross validation is used in our classification model, first validation fold contains the top 10% of the worst-case data n-D points, next validation fold contains the next 10% of the worst-case data and so on.

## 4. Experiments with Interactive Data Classification Approach

The first data set is the Iris data [Dua et al, 2019]. It has 4 dimensions (sepal length, petal length, sepal width and petal width) with a total of 150 cases. The data consists of three classes namely setosa, versicolor and virginica, each class consisting 50 cases. Fig. 9 displays the data in SPCVis software system. The 4 dimensions are denoted as $X_1$, $X_2$, $X_3$ and $X_4$ coordinates.

Class 1 separation is defined by the rule in (4):

$$\text{If } (x_4, x_3) \in R'_1, \text{ then } \mathbf{x} \in \text{class 1.} \tag{4}$$

The optimized coordinate order for separation of classes 1 and 3 are $(X_1, X_3)$ and $(X_2, X_4)$ with $X_3$ and $X_4$ as vertical coordinates. Separation criteria for class 2 and class 3 is given in (5):

$$\text{If } (x_1, x_2, x_3, x_4) \in R_2 \text{ then } \mathbf{x} \in \text{class 2, else } \mathbf{x} \in \text{class 3.} \tag{5}$$

We can further refine rule defined by $R'_2 = R_{21}$ & $R_{22}$ by adding another area $R_3$ to form a new optimized rule in (6):

$$\text{If } (x_1, x_2, x_3, x_4) \in R'_2 \text{ or } R_3, \text{ then } \mathbf{x} \in \text{class 2, else } \mathbf{x} \in \text{class 3.} \tag{6}$$

Fig. 10 visualizes the above rule for separation classes 2 and 3. The accuracy obtained in 10-fold cross validation with worst case split is **100%.**

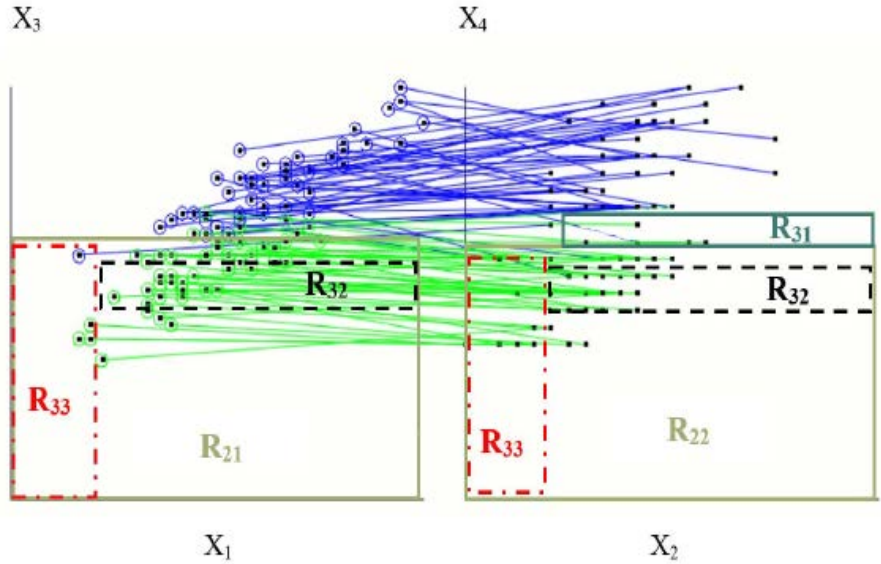

**Fig. 9.** Visualization of rule for R′₁ on Iris dataset (4-D) for class 1 separation.

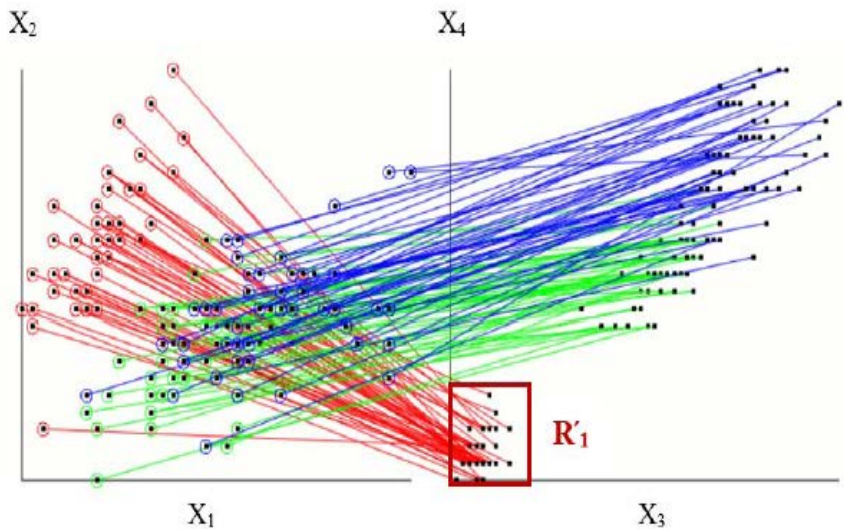

Fig. 10. **Visualization of rule for R₂ and R₃ on Iris dataset (4-D) for classes 2 and 3 separation.**

The second dataset is Wisconsin Breast Cancer (WBC) dataset [Dua et al, 2019] contains 699 cases of data with 9 features. In this dataset, 16 cases were incomplete and hence were removed. Remaining data with 683 cases consists of 444 benign cases and 239 malignant instances. Fig. 3 displays WBC data after loading in SPCVis software system. Fig. 11 visualizes analytical rules for R$_5$ and R$_6$ generated for benign class classification.

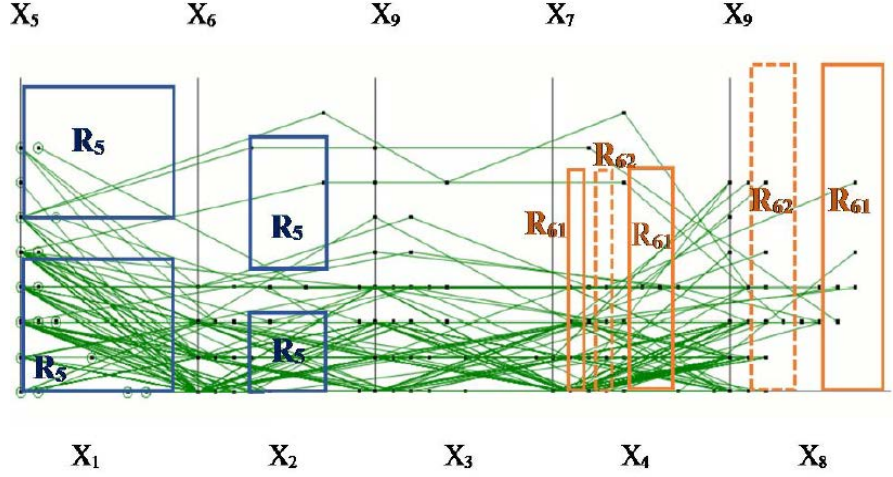

**Fig. 11.** Visualization of rules for R$_5$ and R$_6$ on WBC dataset (9-D).

The analytical rules for R$_5$ and R$_6$ for classification of class 1 is defined in (7).

$$\text{If } (x_1, x_5, x_2, x_6, x_4, x_7, x_8, x_9) \in \text{R}_5 \text{ \& R}_6, \text{ then x} \in \text{class 1} \tag{7}$$

$$\text{R}_6 = \text{R}_{61} \text{ or R}_{62}. \tag{8}$$

The accuracy obtained after 10-fold cross validation technique with worst case heuristics is **99.56%.**

The third dataset consists of seeds data with 7 dimensions and 210 instances. The data contain three classes: Kama, Rosa and Canadian, based on the characteristics of wheat kernels like seed perimeter area, length and width of kernel etc. Each class consists of 70 instances [Dua et al, 2019]. The data are loaded, and the coordinates are reordered to find the prominent class separation along vertical coordinates. The analytical rules are generated based on the vertical separation. Fig. 12 displays seeds data with areas for analytical rules R$_1$ and R$_2$ for class 2 (green) separation. Due to odd number of coordinates X$_2$ coordinate is duplicated as the 8$^{th}$ coordinate. The rule for class 2 classification is defined in (9).

$$\textbf{r}_2\text{: If } (x_1, x_7, x_4, x_2, x_6) \in (\text{R}_1 \text{ or R}_2), \text{ then } \textbf{x} \in \text{class 2} \tag{9}$$

$$\text{R}_2 = \text{R}_{21} \text{ or R}_{22} \tag{10}$$

The accuracy obtained after 10-fold cross validation technique with worst case heuristics is **100%.**

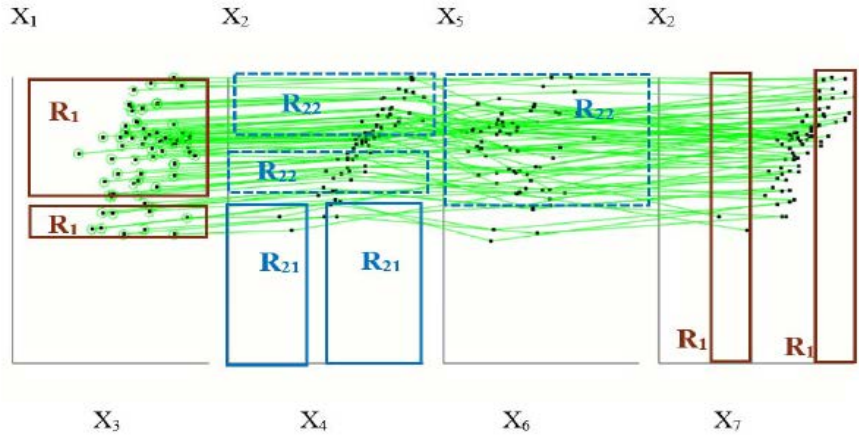

**Fig. 12.** Visualization of rules for $R_1$ and $R_2$ on Seeds dataset (7-D) for class 1 separation with all cases from class 2.

## 5. Data Classification Using Automation

In most of the scenarios, interactive approach to data classification works well for small dataset. When handling large data, this approach becomes tedious and time consuming. Also, due to the display of large number of data cases within limited space would lead to occlusion and it becomes challenging for the end users to find the patterns. To overcome this problem, automation is used with patterns discovered automatically with minimal human intervention. Automation is implemented using: (1) Coordinate Order Optimizer (COO) algorithm, and (2) Genetic algorithm (GA).

These algorithms are used in combination with non-linear scaling, to enhance the data interpretability. Fig. 13 summarizes the automated data classification approach.

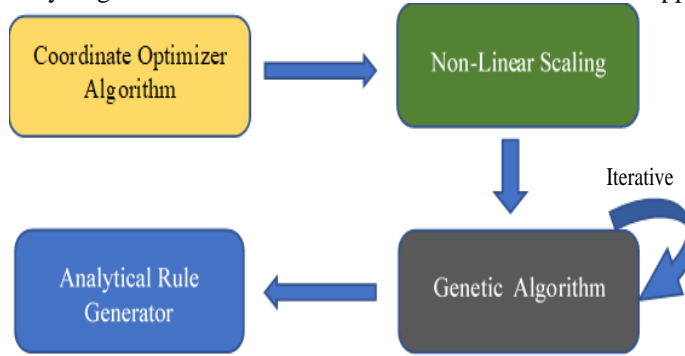

**Fig. 13.** Overview of Automation for Data Classification in SPCVis.

*Coordinate Order Optimizer Algorithm.* It optimizes the order of coordinates by primarily using Coefficient of Variation (CV) [Everitt, 1998] parameter, also called as relative standard deviation (RSD) defined as the ratio of the standard deviation $\sigma$ to the mean $\mu$ of the given data sample,

$$C_v = \frac{\sigma}{\mu} \tag{11}$$

CV is a standardized measure of dispersion of data distribution. It is computed individually per coordinate for each class. Next, the mean of all CVs of classes of each coordinate is calculated. Lesser the mean of CV, lesser the data dispersion along the coordinate. The coordinate with the least mean of CV is considered the *best coordinate*. Hence, the coordinates are arranged in the descending order of the mean CV values. Fig. 14 displays WBC data before and after order of coordinates is optimized, respectively. In the optimized order of coordinates (Fig. 14b), the green class is settled at the bottom and red class on the top, while in the Fig. 14a, more green lines are at the top along with red class and more red lines in the bottom along with green lines.

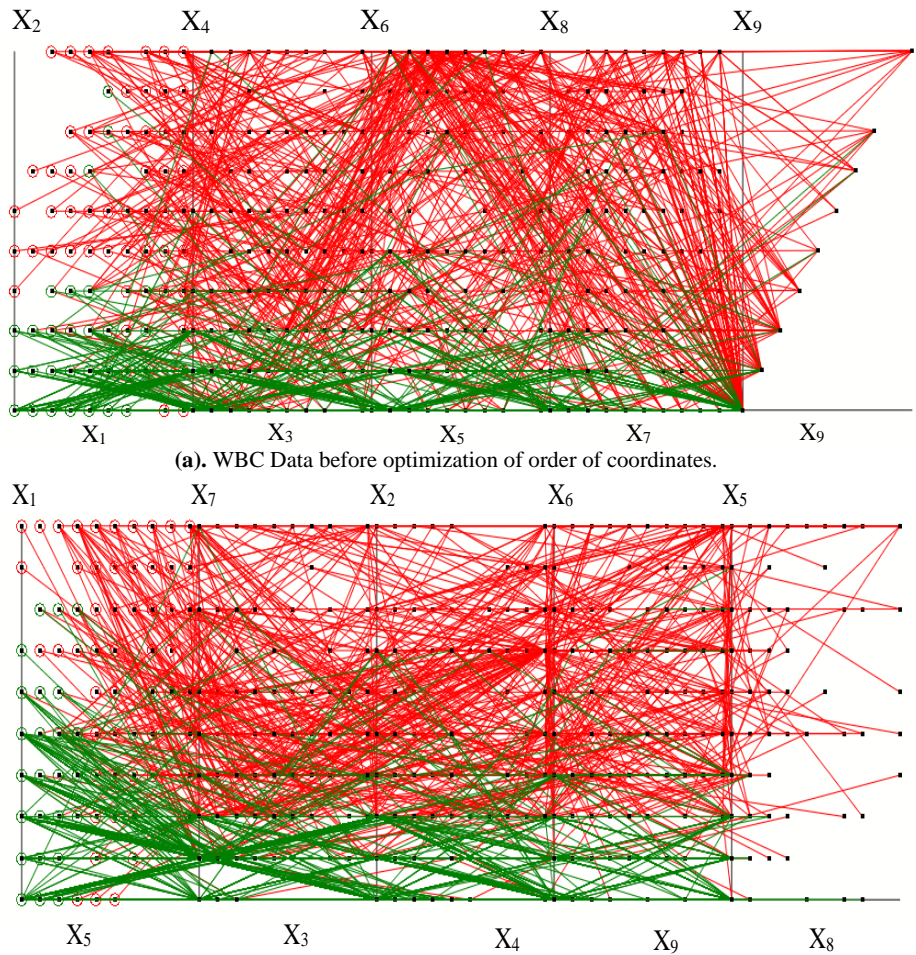

**(a).** WBC Data before optimization of order of coordinates.

**(b).** WBC Data after optimization of order of coordinates (lesser green cases on top).

**Fig. 14.** Visualization of WBC Data before and after applying COO Algorithm.

*Non-Linear Scaling*: The threshold for all the vertical coordinates is calculated from the average of the bottom class taken from vertical coordinates. Non-linear scaling is performed using equation (1) for all the vertical coordinates. This improves data interpretability and provides better visualization of separation of classes. Fig. 4 shows the output of non-linear scaling that enhances the visual separation of classes.

*Genetic Algorithm*: This algorithm is used to generate the optimized areas of high fitness or purity [Cowgill et al, 1999] based on which the analytical rules are created for further classification. An overview of the implementation of Genetic Algorithm in our approach is shown in Fig. 15 for discovering the areas for classification. In this context, areas are referred as Area of Interest (AOI).

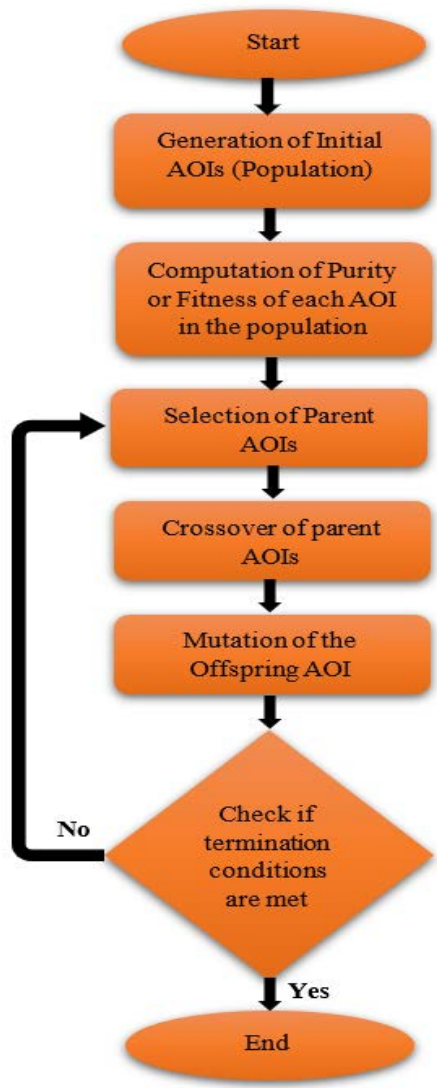

**Fig. 15.** Genetic algorithm flow chart used in SPCVis data classification.

*Initial Population Selection*: The initial population contains randomly generated rectangles (see Fig. 16) defined by the fixed ratio *r* of each coordinates. For instance, *r* can be 0.1 of length of the coordinate. The data are normalized to [0,1] interval. The generation of the Area of Interest (AOI) in the SPCVis using genetic algorithm is an iterative process where each iteration creates a generation [Banzhaf et al, 1998] of a new set of AOIs. Before generating the AOIs, a search space [Bouali et al, 2020] is defined containing maximum cases belonging to the class on which we build analytical rules. Consider a situation where analytical rules are being built to classify class *k*. Let $x_{imax}(k)$ and $x_{imin}(k)$ be the maximum and minimum data points respectively belonging to coordinate $X_i$ and $x_{jmax}(k)$ and $x_{jmin}(k)$ be the maximum and minimum data points respectively belonging to coordinate $X_j$. The number of Areas of Interests $N_{AOI}$ generated to classify class *k* within a coordinate pair $(X_i, X_j)$ is given in (12) and (13),

$$N_{AOI} = \frac{S(k)}{r^2} \tag{12}$$

$$S(k) = (x_{imax}(k) - x_{imin}(k)) \, (x_{jmax}(k) - x_{imin}(k)) \tag{13}$$

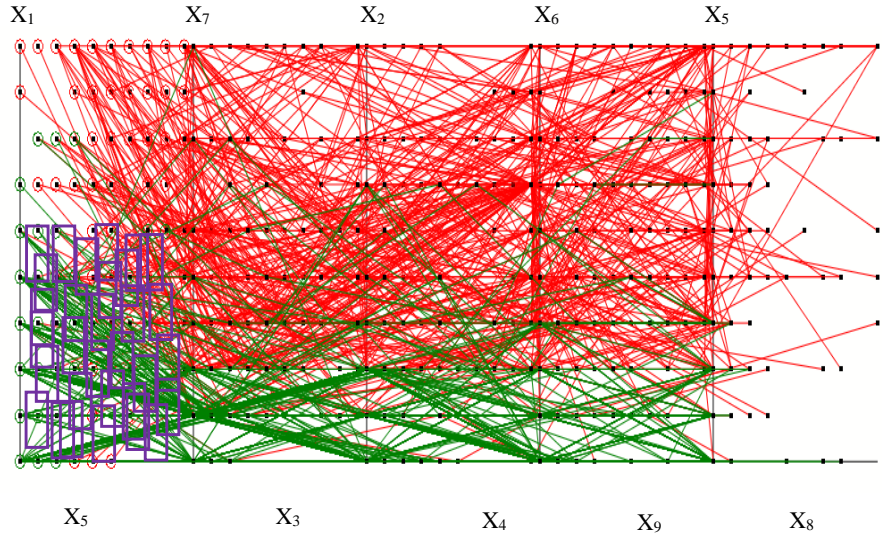

**Fig. 16.** Random generation of areas in WBC data.

*Parents Selection*: This stage involves selection of the AOIs or parents to generate a AOI called the offspring, i.e., combing two areas to form a bigger area. The parents are selected based on two criteria (1) *purity* or fitness, and (2) *proximity*.

*Purity* or Fitness of an AOI with respect to class *k* for a given pair of coordinates $(X_i, X_j)$ is defined as the ratio of the number of data points belonging to class *k* to the total number of data points within the AOI. Let $AOI_t$ be an AOI in the coordinate pair $(X_i, X_j)$. The purity $P_k(AOI_t)$ of a single AOI with respect to class *k* is as follows,

$$P_k(AOI_t) = \frac{N_k(AOI_t)}{N(AOI_t)} \tag{14}$$

where, $N_k(AOI_t)$ is the number of points $(x_i, x_j)$ in $AOI_t$ in $(X_i, X_j)$ that belong to lines from class $C_k$,

$$N_k(AOI_t) = \|\{(x_i, \ x_j): (x_i, x_j) \in AOI_t \& C_k\}\| \tag{15}$$

and $N(AOI_t)$ is the total number of points $(x_i, x_j)$ within a given $AOI_t$ in $(X_i, X_j)$,

$$N(AOI_t) = \|\{(x_i, x_j): (x_i, x_j) \in AOI_t\}\| \tag{16}$$

After computing the purity of the AOIs, the parents with closest *Proximity* (nearest parents) are selected for the next stage, i.e., the crossover. While proximity can be defined in multiple ways, in our experiments we applied the Euclidian distance between the mid-point of two areas within the given coordinate pair commonly used in Genetic algorithms for similar tasks [Bouali et al, 2020]. For instance, given the mid-point of two AOIs $\{(x_{im1}, x_{jm1}), (x_{im2}, x_{jm2})\}$ within coordinate pair $(X_i, X_j)$, the proximity of two AOIs is given in (17):

$$Proximity = \sqrt{(x_{im1} - x_{im2})^2 + (x_{jm1} - x_{jm2})^2} \tag{17}$$

*Crossover*: Once the parents with highest *Purity* and closest *Proximity* (nearness) are selected, the parent AOIs are combined to form a new AOI (offspring). A single parent $AOI_{Pg}$ is represented below:

$$AOI_{Pg} = P_g(x_1, x_2, y_1, y_2) \tag{18}$$

Here $x_1, x_2, y_1, y_2$ are left, right, bottom ant top coordinates of the rectangular $AOI_{Pg}$ and is represented in the Fig. 17.

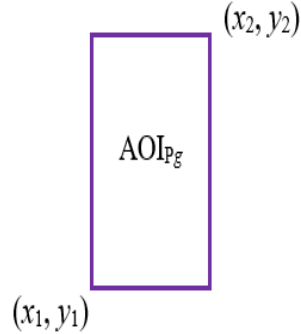

**Fig. 17**. Representation of single parent $AOI_{Pg}$ from generation $g$.

A crossover of two parents $AOI_{P1g}$ and $AOI_{P2g}$ from generation $g$ to produce an offspring $AOI_{O1g}$ within a given coordinate pair $(X_i, X_j)$ is represented as:

$$AOI_{O1g} = F(AOI_{P1g}, AOI_{P2g}) \tag{19}$$

The two parents are defined as:

$$\text{AOI}_{\text{P1}g} = \text{P}_{1g}(x_{11}, x_{12}, y_{11}, y_{12}) \tag{20}$$

$$\text{AOI}_{\text{P2}g} = \text{P}_{2g}(x_{21}, x_{22}, y_{21}, y_{22}) \tag{21}$$

The function $F(\text{AOI}_{\text{P1}g}, \text{AOI}_{\text{P2}g})$ is defined in (22) as an envelope around these two AOIs.

$$\text{F}(\text{AOI}_{\text{P1}g}, \text{AOI}_{\text{P2}g}) = \{\min(x_{11}, x_{21}), \max(x_{12}, x_{22}), \min(y_{11}, y_{21}), \max(y_{12}, y_{22})\} \tag{22}$$

Fig. 18 displays different types of crossovers of the parent AOIs (with and without overlapping, or diagonally overlapping).

*Mutation*: In genetic algorithm, certain characteristics of the offspring generated from the previous generation are modified (mutated) in order to speed up the process of reaching an optimized solution. This includes modifying the characteristics of the offspring either by flipping, swapping or shuffling the properties that represent the offspring. For instance, if the offspring is represented by bits, then the mutation by flipping would include switching some of the bits from 0 to 1 or vice versa [Rifki, et al, 2012].

Since the objective of mutation is to generate an offspring with better characteristics than its parents, in our proposed technique, we generate the mutated offspring by interactively *generating a new parent AOI* with high *Purity* and close *Proximity* with the automatically generated parent AOI. This results in an offspring with better characteristics compared to its parents in terms of size and purity. Fig. 19a represents the automatically generated parent AOI in purple (straight line) and interactively generated parent AOI (dotted lines). The resulting mutated offspring AOI has superior characteristics compared to its parents with high purity and larger area compared to its previous generation as displayed in Fig. 19b.

*Termination*: Since genetic algorithm process is iterative, there are several conditions based on which the process can be terminated. In our proposed method, we use two techniques as the termination criteria: (1) areas with highest fitness or purity (100%) are generated, or (2) manual inspection termination in SPCVis. If either of the two criteria is met, the process is terminated.

*Analytical Rule Generator*: This is similar to the second step performed in IVLC algorithm as discussed in section 3 of this chapter. The only change that the areas used here are generated from genetic algorithm whereas in section 3, the areas used are interactively generated.

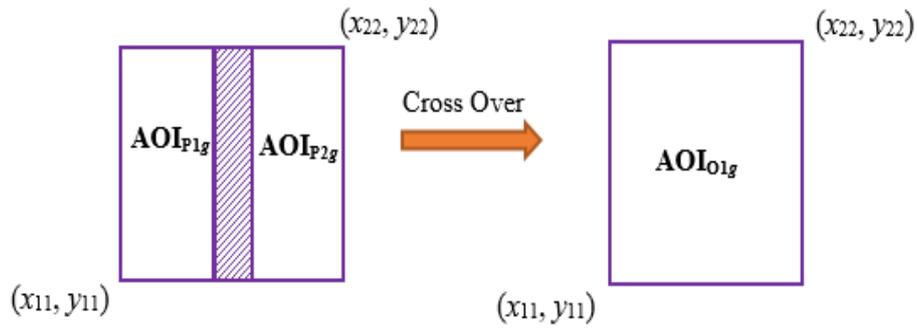

**(a).** Cross Over of two overlapping parent AOIs.

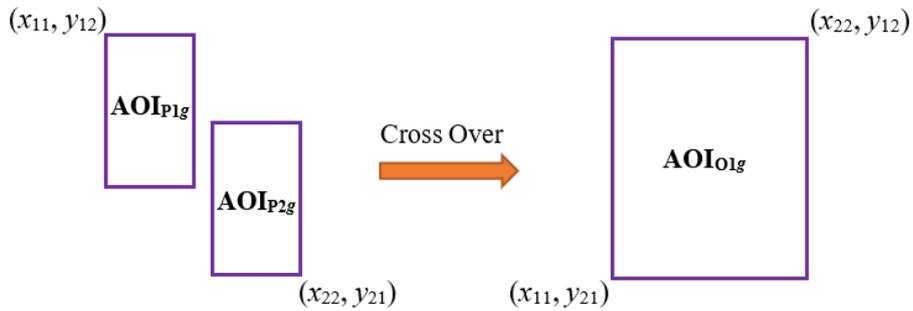

**(b).** Cross Over of two non - overlapping parent AOIs.

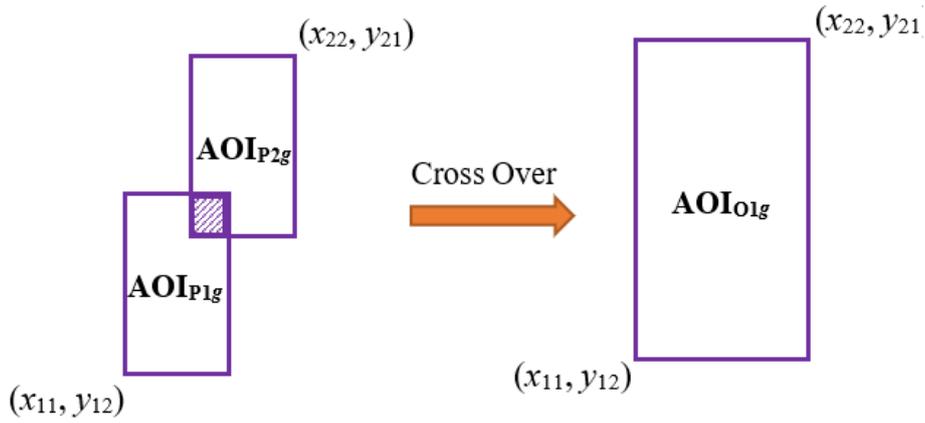

**(c).** Cross Over of two diagonally overlapping parent AOIs.

**Fig. 18**. Different types of Cross Overs of two parent AOIs to generate Offspring AOI.

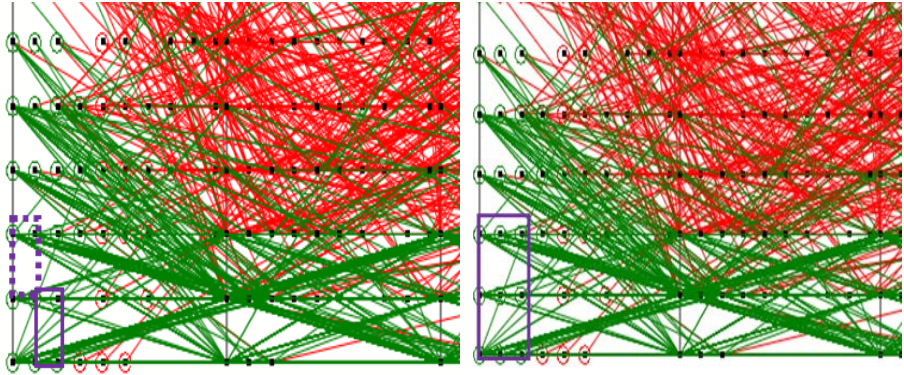

**(a).** Parent AOI (dotted lines) generated automatically (straight lines) and interactively (dotted lines) with high purity in generation *g*.

**(b).** Mutated Offspring in generation *g* + 1.

**Fig. 19.** Visualizations of consecutive generations of AOIs in WBC data.

## 6. Experiments with Automated Data Classification Approach

Experiments are conducted with same data sets mentioned in section 5 i.e., WBC, Iris and Seeds datasets. In addition to these data sets, experiments are also conducted on Air pressure system failure at Scania trucks. This dataset consists of 60,000 cases with 170 features. Compared to interactive approach, the automated techniques provided better results with smaller number of areas and less iterations.

The experiment is conducted on iris dataset, as discussed in section 5. Class 1 is classified after running the COO algorithm and genetic algorithm. The optimized order of the coordinates for class 1 classification is $(X_4, X_3)$ and $(X_1, X_2)$. It is displayed in Fig. 20. The rule for class 1 (green) separation is:

$$\mathbf{r}_1: \text{If } (x_4, x_3) \in \text{ R}_{11}, \text{ then } \mathbf{x} \in \text{ class 1} \tag{23}$$

After class 1 separation, Coordinate Order Optimizer is run again for classifying the remaining classes. The optimized order of coordinates for class 2 and class 3 separation are $(X_1, X_3)$ and $(X_2, X_4)$. The visualization of classes 2 and 3 after reordering the coordinates is shown in Fig. 21.

Fig. 21 clearly displays separation of classes 2 and 3 along the vertical coordinates. This visualization is further enhanced by applying the non-linear scaling with following thresholds on coordinates: 0.7 on $X_3$ and 0.71 on $X_4$.

Genetic algorithm is run on class 2 and class 3 data to generate the areas. Visualization of non-linear scaling along with the areas are displayed in Fig. 22a with

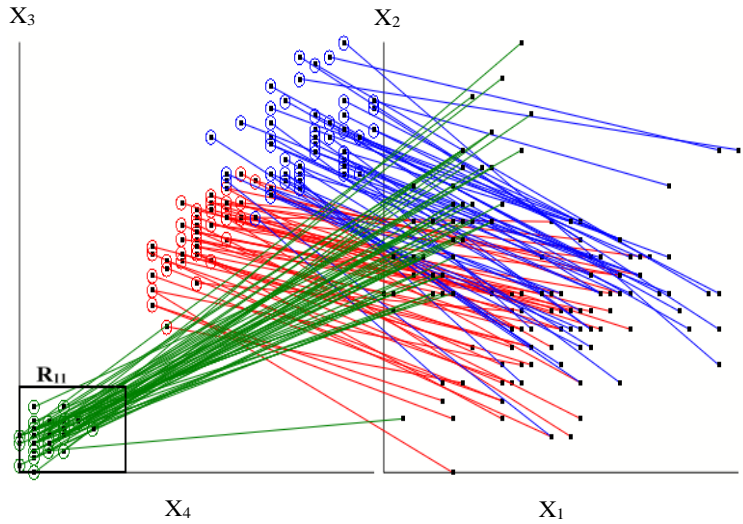

**Fig. 20.** Visualization of Iris data with class 1 separation rule.

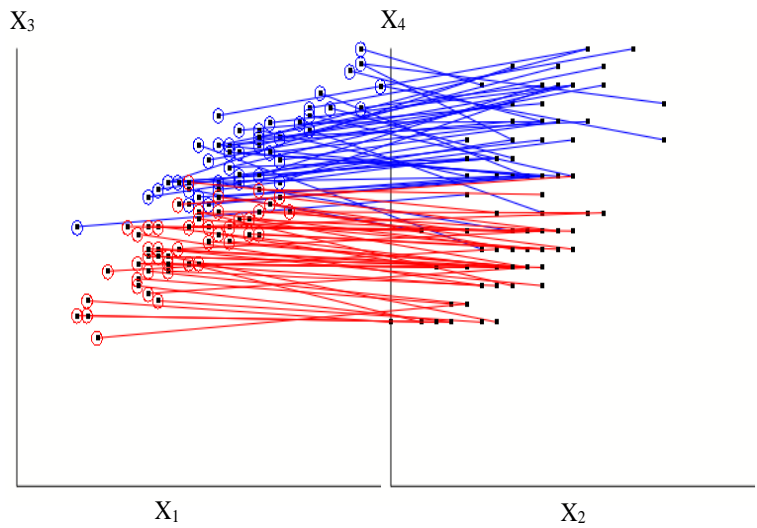

**Fig. 21.** Visualization of Iris data with classes 2 and 3 after reordering the coordinates.

10 instances. Fig. 22b displays the same visualization with all the cases from class 2 and class 3.

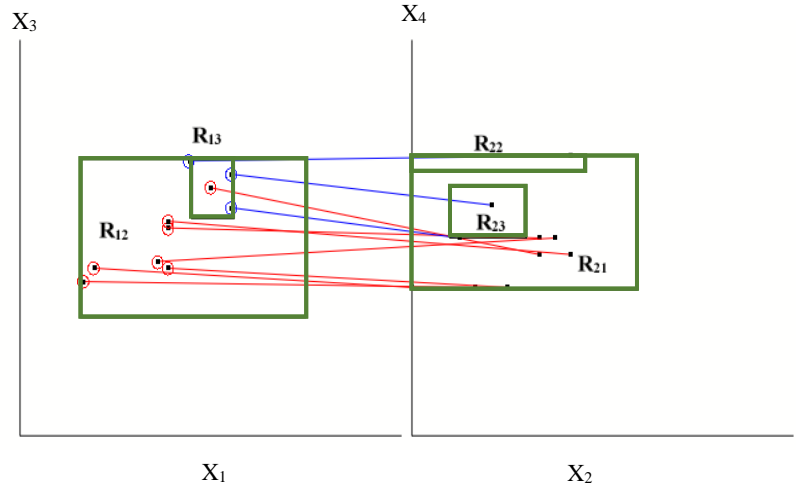

**(a).** Visualization of rule $\mathbf{r_2}$ on Iris dataset for classes 2 and 3 separation with 10 cases.

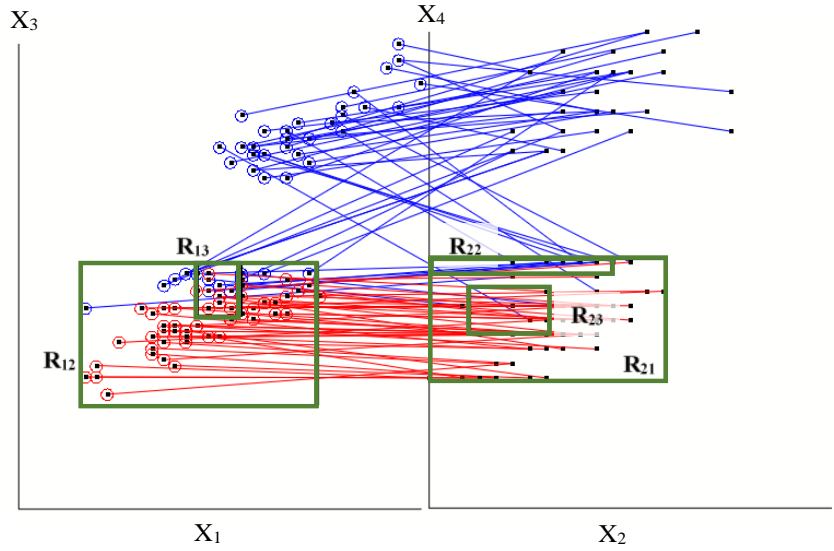

**(b).** Visualization of rule $\mathbf{r_2}$ on Iris dataset for classes 2 and 3 separation with all the cases.

**Fig. 22.** Visualization of rule $\mathbf{r_2}$ on Iris dataset for classes 2 and 3 separation.

The area $R_2$ for classes 2 and 3 classification is:

$$R_2 = R_{12} \ \& \ R_{21} \ \& \ \neg R_{13} \ \& \ \neg R_{23} \ \& \ \neg R_{22} \tag{24}$$

The rule $\mathbf{r_2}$ for class 2 classification for Iris data is defined below:

$$\mathbf{r_2}: \text{If } (x_1, x_2, x_3, x_4) \in \ R_2 \text{ , then } \mathbf{x} \in \text{ class 2} \tag{25}$$

The rule $\mathbf{r_3}$ for class 3 classification for Iris data is defined below:

$$\mathbf{r_3}: \text{If } (x_1, x_2, x_3, x_4) \in \text{ R}_3 \text{ , then } \mathbf{x} \in \text{ class 3} \tag{26}$$

where

$$R_3 = (\neg R_{11} \ \& \ \neg R_2) \tag{27}$$

The area parameters generated for Iris data classification is listed in Table 2. The accuracy obtained for Iris data classification with 10-fold cross validation using worst-case heuristics approach is **100%.**

Table 2. Parameters of the areas generated for Iris data classification.

| Rectangle | Left | Right | Bottom | Top | Coordinate Pair |
|-----------|------|-------|--------|-----|-----------------|
| $R_{11}$ | 0.0 | 0.3 | 0.0 | 0.2 | $(X_1, X_3)$ |
| $R_{12}$ | 0.16 | 0.75 | 0.3 | 0.7 | $(X_1, X_3)$ |
| $R_{13}$ | 0.45 | 0.56 | 0.55 | 0.7 | $(X_1, X_3)$ |
| $R_{21}$ | 0.0 | 0.59 | 0.37 | 0.71 | $(X_2, X_4)$ |
| $R_{22}$ | 0.0 | 0.45 | 0.67 | 0.71 | $(X_2, X_4)$ |
| $R_{23}$ | 0.1 | 0.3 | 0.5 | 0.63 | $(X_2, X_4)$ |

The second dataset is WBC dataset, as discussed in section 5. Running the COO algorithm produced the following order of coordinates: $(X_5, X_1)$, $(X_3, X_7)$, $(X_4, X_2)$, $(X_9, X_6)$ and $(X_8, X_5)$. Fig. 23a displays WBC data visualized in SPCVis with 12 cases of class 1 data along with the areas.

The rectangle $R_{km}$ is $m^{\text{th}}$ rectangle in the $k^{\text{th}}$ pair of coordinates. For instance, in Fig. 23a, rectangle $R_{24}$ is a $4^{\text{th}}$ rectangle in the second pair of coordinates that is $(X_3, X_7)$. Fig. 23b displays all the cases from both classes of WBC data along with the non-linear scaling with following thresholds on coordinates: 0.6 on $X_1$, 0.25 on $X_7$ and $X_2$ and 0.3 on $X_6$.

The areas $R_1$-$R_3$ are defined as follows:

$$R_1 = R_{11} \& \neg R_{14} \ \& \ R_{41} \& \neg R_{42} \& \ (\neg R_{31} \text{ or } \neg R_{23} \text{ or } \neg R_{32}) \tag{28}$$

$$R_2 = R_{12} \& \neg R_{15} \ \& \ (R_{21} \text{ or } R_{24} \text{ or } R_{33} ) \tag{29}$$

$$R_3 = R_{13} \ \& \ R_{24} \tag{30}$$

The rule $\mathbf{r}$ is defined below:

$$\begin{aligned} \mathbf{r} = \text{If } (x_1, x_2, x_3, x_4, x_5, x_6, x_7, x_9) \ \in \ R_1 \text{ or } R_2 \text{ or } R_3 \text{ then} \\ \mathbf{x} \in \text{ Class 1, else } \mathbf{x} \in \text{ Class 2} \end{aligned} \tag{31}$$

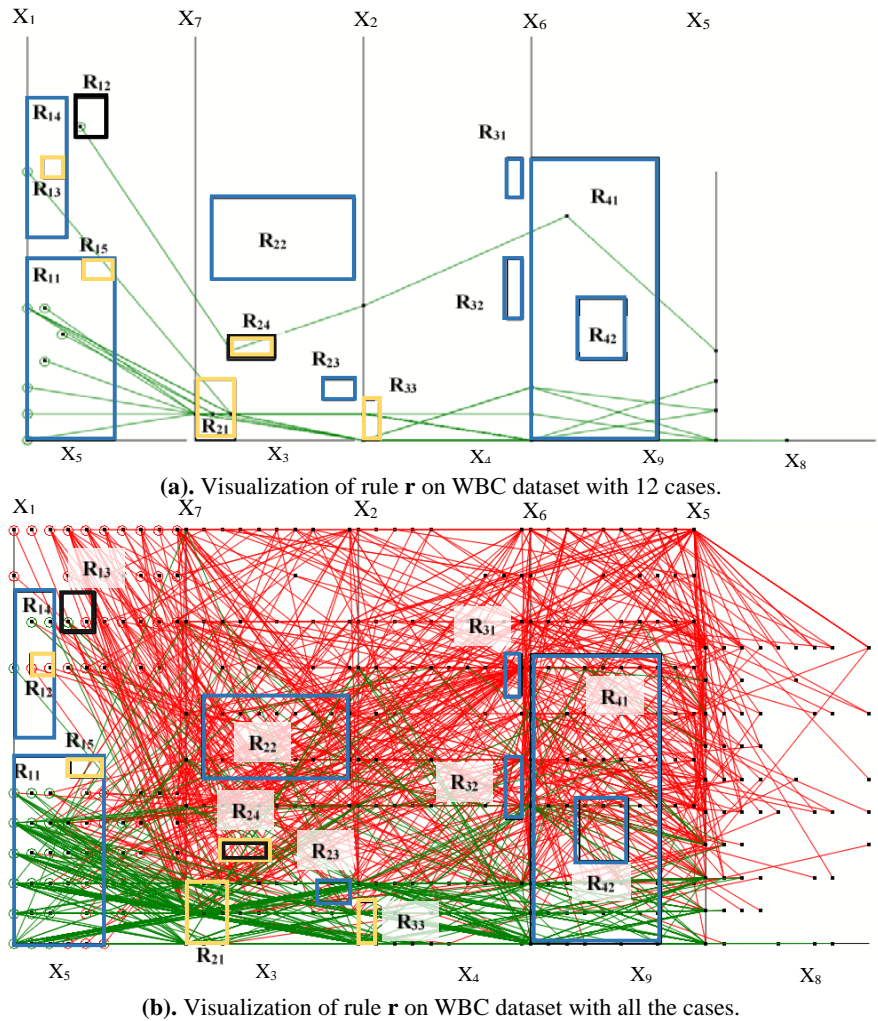

**(a).** Visualization of rule **r** on WBC dataset with 12 cases.

**(b).** Visualization of rule **r** on WBC dataset with all the cases.

**Fig. 23.** Visualization of rule **r** on WBC dataset for class 1 separation.

The area coordinates are given in the Table 3. The accuracy obtained after 10-fold cross validation technique with worst case heuristics is **99.71%.**

Seeds data, as discussed in section 5 consists of 7 dimensions. Due to odd number of dimensions, $X_7$ is duplicated to display the data in SPCVis (see Fig. 24).

Analytical rules for class 3 (blue) are generated after running the COO algorithm and genetic algorithm. The optimized order of the coordinates for class 1 classification is $(X_3, X_1)$, $(X_7, X_5)$, $(X_6, X_2)$ and $(X_3, X_4)$. Fig. 25 represents the optimized order of coordinates.

Table 3. Parameters of the rectangles generated for WBC data classification.

| Rectangle | Left | Right | Bottom | Top | Coordinate Pair | |
|---|---|---|---|---|---|---|
| $R_{11}$ | 0.0 | 0.55 | 0.0 | 0.45 | $(X_5, X_1)$ | |
| $R_{12}$ | 0.3 | 0.5 | 0.75 | 0.85 | $(X_5, X_1)$ | |
| $R_{13}$ | 0.1 | 0.25 | 0.65 | 0.7 | $(X_5, X_1)$ | |
| $R_{14}$ | 0.0 | 0.25 | 0.5 | 0.85 | $(X_5, X_1)$ | |
| $R_{15}$ | 0.40 | 0.55 | 0.40 | 0.45 | $(X_5, X_1)$ | |
| $R_{21}$ | 0.0 | 0.25 | 0.0 | 0.15 | $(X_3, X_7)$ | |
| $R_{22}$ | 0.1 | 1.0 | 0.4 | 0.6 | $(X_3, X_7)$ | |
| $R_{23}$ | 0.8 | 1.0 | 0.1 | 0.15 | $(X_3, X_7)$ | |
| $R_{24}$ | 0.2 | 0.5 | 0.2 | 0.25 | $(X_3, X_7)$ | |
| $R_{31}$ | 0.9 | 1.0 | 0.6 | 0.7 | $(X_4, X_2)$ | |
| $R_{32}$ | 0.9 | 1.0 | 0.3 | 0.45 | $(X_4, X_2)$ | |
| $R_{33}$ | 0.0 | 0.1 | 0.0 | 0.12 | $(X_4, X_2)$ | |
| $R_{41}$ | 0.0 | 0.8 | 0.0 | 0.7 | $(X_9, X_6)$ | |
| $R_{42}$ | 0.3 | 0.6 | 0.2 | 0.35 | $(X_9, X_6)$ | |

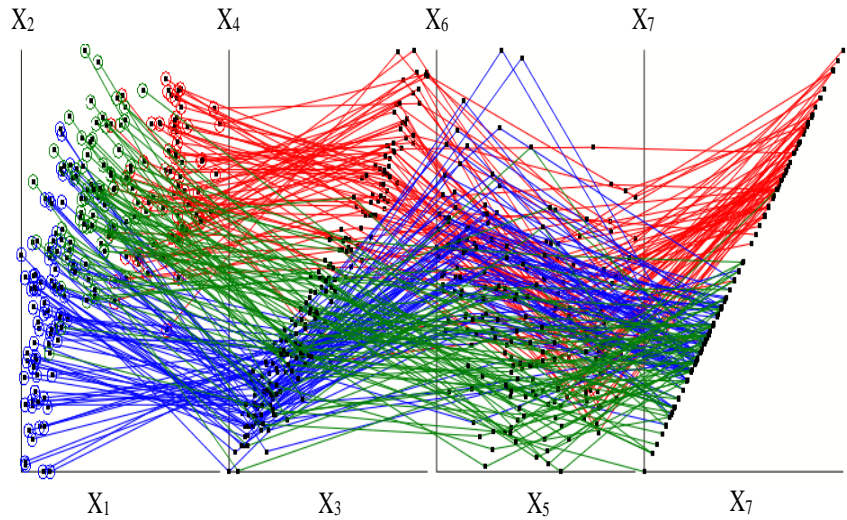

**Fig. 24.** Visualization of Seeds dataset (7-D) with all the three classes in SPCVis.

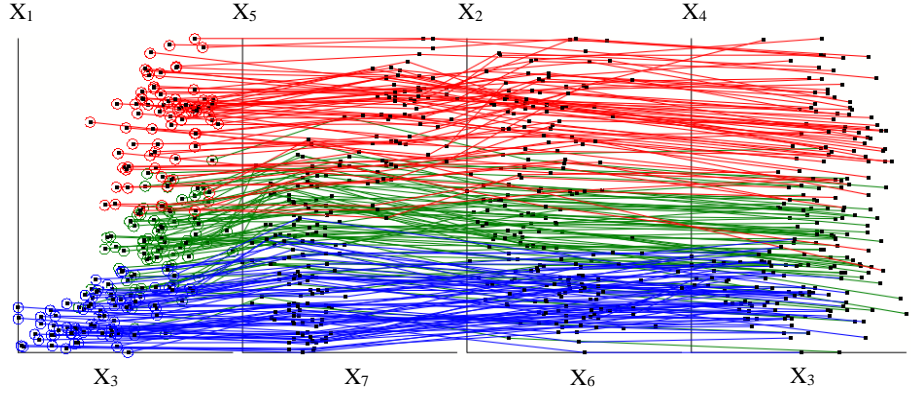

**Fig. 25.** Visualization of Seeds dataset (7-D) with all the three classes in SPCVis after coordinate order optimization.

Non- linear scaling is then performed with following thresholds on coordinates: 0.4 on $X_1$, 0.45 on $X_5$, 0.4 on $X_2$ and 0.4 on $X_4$. The visualization of classes 2 and 3 after reordering the coordinates and non-linear scaling is shown in Fig. 26.

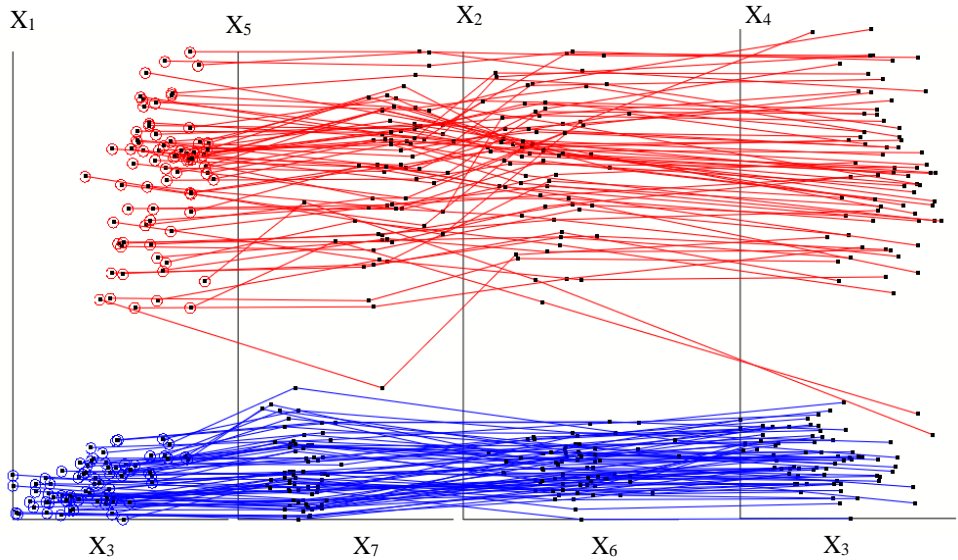

**Fig. 26.** Visualization of Seeds dataset with classes 2 (red) and 3 (blue) after performing non-linear scaling on optimized order of coordinates.

Areas are generated by running genetic algorithm and analytical rules are built using these generated areas. Fig. 27a visualizes all the three classes with non-linear scaling and the areas for class 3 (blue) separation. The area is defined below:

$$R_1 = R_{11} \And R_{21} \And R_{41} \And (\neg R_{42} \text{ or } \neg R_{31}) \And (\neg R_{12} \And \neg R_{22} \And (\neg R_{32} \text{ or } \neg R_{43})) \quad (32)$$

The rule **$r_1$** for class 3 classification is given as:

$$\mathbf{r_1}: \text{If } (x_1, x_2, x_3, x_4, x_5, x_6, x_7) \in R_1, \text{ then } \mathbf{x} \in \text{ class 3} \qquad (33)$$

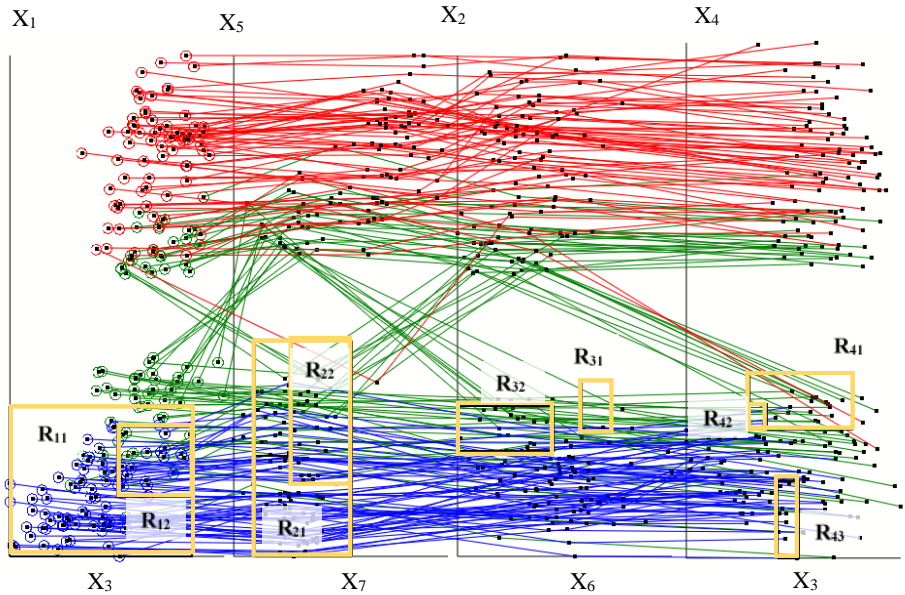

**(a).** Cases covered by rule $\mathbf{r_1}$ on Seeds dataset for class 3 (blue) separation.

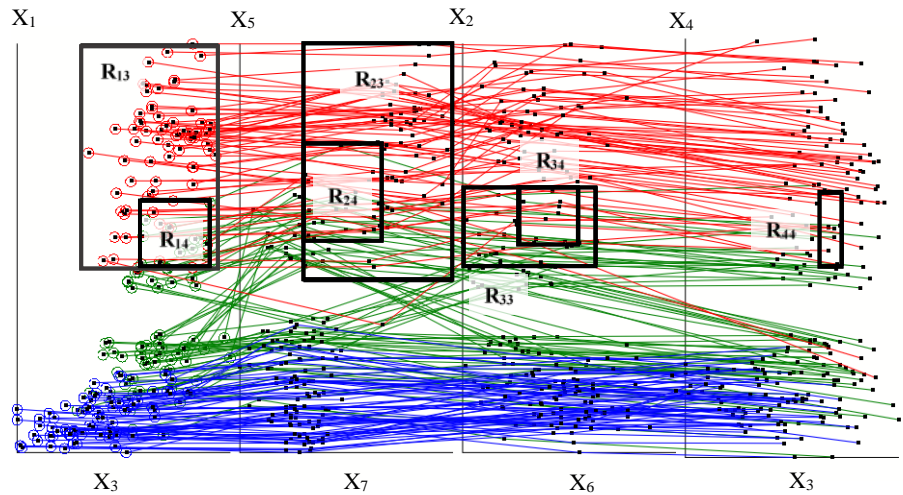

**(b).** Cases covered by rule $\mathbf{r_2}$ on Seeds dataset for class 2 (red) separation.

**Fig. 27.** Visualization of rules $\mathbf{r_1}$ and $\mathbf{r_2}$ on Seeds dataset for classes 2 and 3 separation with all the cases.

The parameters of the areas generated by genetic algorithm are listed in Table 4.

Table 4. Parameters of the areas generated for classification in Seeds data in 1st iteration.

| Rectangle | Left | Right | Bottom | Top | Coordinate Pair |
|-----------|------|-------|--------|-----|-----------------|
| $R_{11}$ | 0.0 | 0.85 | 0.0 | 0.3 | $(X_3, X_1)$ |
| $R_{12}$ | 0.5 | 0.85 | 0.12 | 0.26 | $(X_3, X_1)$ |
| $R_{21}$ | 0.1 | 0.55 | 0.0 | 0.43 | $(X_7, X_5)$ |
| $R_{22}$ | 0.27 | 0.55 | 0.2 | 0.43 | $(X_7, X_5)$ |
| $R_{31}$ | 0.58 | 0.72 | 0.25 | 0.35 | $(X_6, X_2)$ |
| $R_{32}$ | 0.0 | 0.45 | 0.21 | 0.3 | $(X_6, X_2)$ |
| $R_{41}$ | 0.32 | 0.8 | 0.25 | 0.36 | $(X_3, X_4)$ |
| $R_{42}$ | 0.32 | 0.4 | 0.25 | 0.3 | $(X_3, X_4)$ |
| $R_{43}$ | 0.45 | 0.55 | 0.0 | 0.15 | $(X_3, X_4)$ |

The cases that do not follow the rule generated for class 3 is sent to the next iteration where rules are generated for other classes. Areas are generated again by running genetic algorithm and analytical rules are built using these generated areas. The visualization with all the three cases along with non-linear scaling and the areas are displayed in Fig. 27b. In this case the rule is generated for class 2 (red). The optimized order of the coordinates remains the same. Non- linear scaling with the same threshold as in the previous iteration is performed on the vertical coordinates. The rule for class 2 (red) separation is:

$$R_2 = R_{13} \,\&\, R_{23} \,\&\, ((\neg R_{14} \,\&\, \neg R_{24} \,\&\, \neg R_{33}) \,\&\, (\neg R_{34} \,\&\, \neg R_{44})) \tag{34}$$

$$R_3 = (\neg R_1 \,\&\, \neg R_2) \tag{35}$$

The rule **r₂** for class 2 separation is given below:

$$\mathbf{r_2}\text{: If } (x_1, x_2, x_3, x_4, x_5, x_6, x_7) \in R_1, \text{ then } \mathbf{x} \in \text{ class 2} \tag{36}$$

The rule **r₃** for class 1 separation is given below:

$$\mathbf{r_3}\text{: If } (x_1, x_2, x_3, x_4, x_5, x_6, x_7) \in R_3, \text{ then } \mathbf{x} \in \text{ class 1} \tag{37}$$

The accuracy obtained for Seeds data classification with this approach and applying 10-fold cross validation using worst-case heuristics validation split is **100%.** The parameters of the areas generated by genetic algorithm are listed in Table 5.

Table 5. Parameters of the areas generated for classification in Seeds data in 2nd iteration.

| Rectangle | Left | Right | Bottom | Top | Coordinate Pair |
|-----------|------|-------|--------|-----|-----------------|
| $R_{13}$ | 0.3 | 0.94 | 0.45 | 1.0 | $(X_3, X_1)$ |
| $R_{14}$ | 0.58 | 0.91 | 0.45 | 0.62 | $(X_3, X_1)$ |
| $R_{23}$ | 0.3 | 1.0 | 0.42 | 1.0 | $(X_7, X_5)$ |
| $R_{24}$ | 0.3 | 0.67 | 0.52 | 0.76 | $(X_7, X_5)$ |
| $R_{33}$ | 0.0 | 0.63 | 0.46 | 0.65 | $(X_6, X_2)$ |
| $R_{34}$ | 0.27 | 0.55 | 0.51 | 0.65 | $(X_6, X_2)$ |
| $R_{44}$ | 0.63 | 0.74 | 0.46 | 0.63 | $(X_3, X_4)$ |

APS (Air Pressure System) failure for Scania Trucks from UCI Repository [Dua et al, 2019] consists of two classes. Class 1 corresponds to the failure in Scania trucks that is not due to the air pressure system and class 2 corresponds to the failure in Scania trucks that is due to the air pressure system. This data consists of 60000 cases and 170 dimensions.

However, the data set contains a large number of missing values. These missing values are replaced by calculating 10% of the maximum value of the corresponding column and multiplying the result by -1. Also, there are 4 columns of data with all 0 values. After imputation and removing the columns without any information, the final data contains 60000 cases with 166 dimensions. As discussed in section 2, visualizing 166 dimensions becomes very challenging. Hence, we use Serpent Coordinate System (SCS) to visualize high dimension data as shown in Fig. 7.

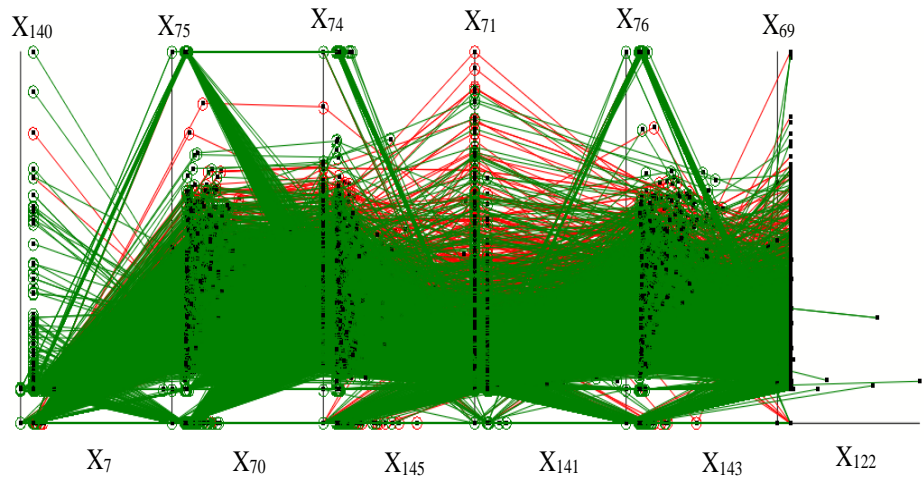

**(a).** APS data with green class on top.

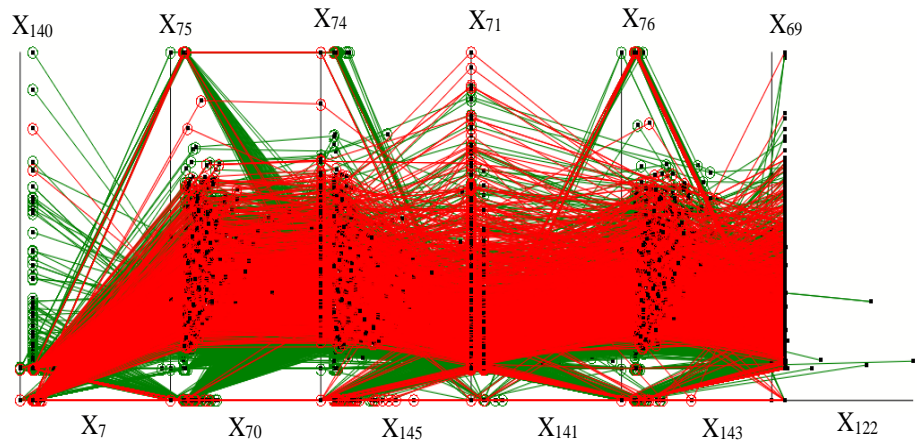

**(b).** APS data with red class on top.

**Fig. 28.** Visualization of 12 best coordinates of APS data in SPCVis.

Data classification using interactive approach becomes very tedious due large number of dimensions and instances. Hence, for this dataset, classification is performed using automation technique. After running the COO algorithm, the result contains 166 coordinates arranged from the most to least optimized coordinates. Here, we start with extracting top 4 coordinates and perform analysis on the data with 2 pairs of coordinates displayed in SPCVis.

The coordinates are gradually increased by pairs until we get the desired results. In this case, 12 coordinates were selected for further analysis. They are $X_7$, $X_{140}$, $X_{70}$, $X_{75}$, $X_{145}$, $X_{74}$, $X_{141}$, $X_{71}$, $X_{143}$, $X_{76}$, $X_{122}$, and $X_{69}$. APS dataset with top 12 coordinates is displayed with green class on top is displayed in Fig. 28a and red class on top is displayed in Fig. 28b.

The visualizations in Fig. 28 display high degree of occlusion even with the best order coordinates. Although there is fair amount of separation observed in the first pair of coordinates. Data in the remaining five pairs are highly occluded. Since there is no clear vertical separation between red class and green class, non-linear scaling becomes insignificant and hence not performed on this data set. However, genetic algorithm is still run on this data to generate areas with high purity. The resulting visualization after running genetic algorithm is displayed in Fig. 29.

Due to two main reasons, the data pattern cannot be interpreted by the end users in this situation: (1) high density of data within the areas generated and (2) small size of the areas generated by genetic algorithm.

To address these issues, we use zooming and averaging. Zoom interactive features wherein the small areas data can be zoomed to view the data more clearly. Fig. 30 displays the zoomed image of are $R_{31}$.

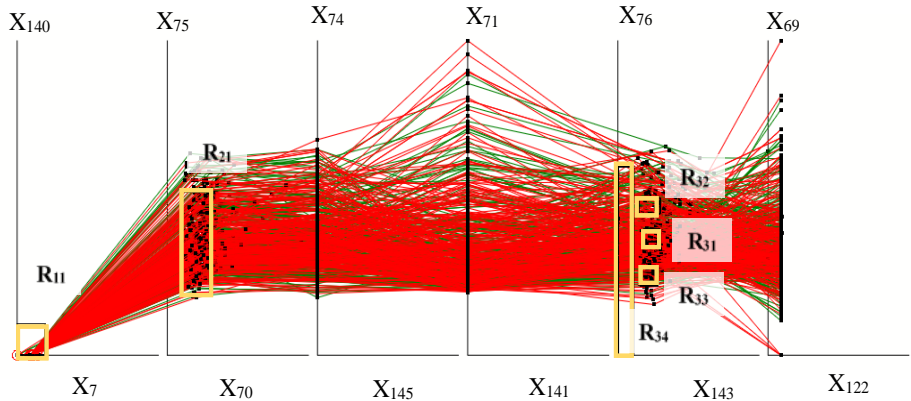

**Fig. 29.** Visualization of APS data with areas generated by Genetic Algorithm for red class classification.

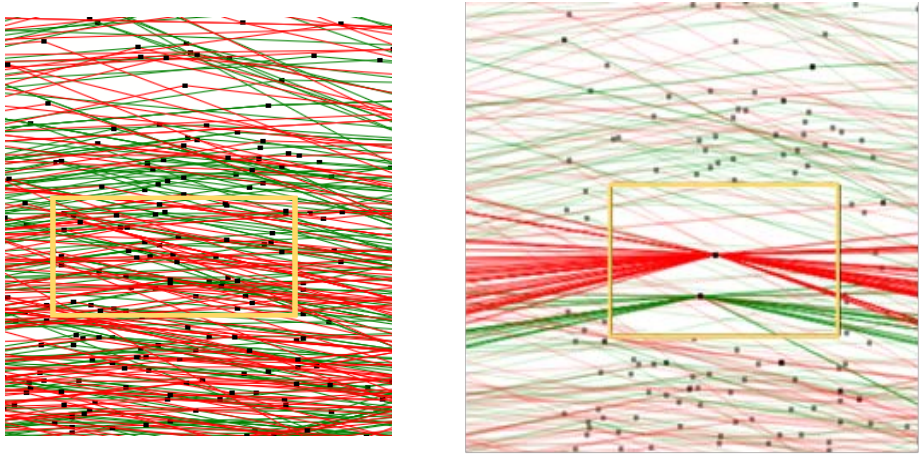

**(a).** Visualization of zoomed $R_{31}$ area in the APS failure data without averaging.

**(b).** Visualization of zoomed $R_{31}$ area in the APS failure data after averaging.

**Fig. 30.** Visualization of $R_{31}$ area in the APS failure data with zooming and averaging.

The zoomed visualization in Fig. 30a solves the problem partially. Although, the data are distinctly visible, the pattern is still hidden. To view the overall distribution of red and green class data, the average of individual class within the area is performed. Figs. 30b and 31 display the averaged red and green class with $R_{31}$ area.

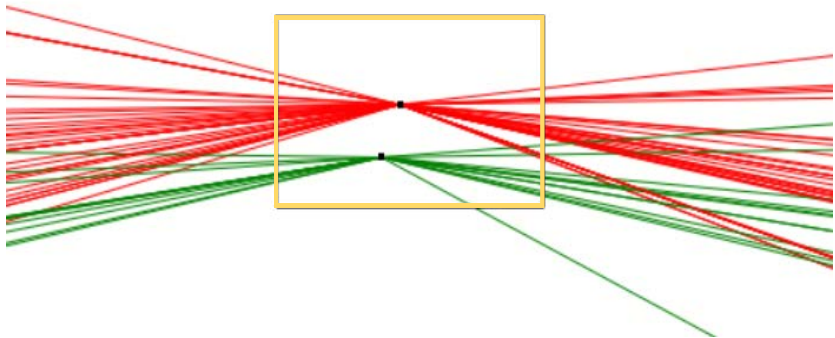

**Fig. 31.** Visualization of zoomed $R_{31}$ area in the APS failure data with averaged classes with the area (without the surrounding data).

Averaging is performed on all the areas generated by the algorithm. Since the areas are generated in the first, second and fifth pair of coordinates, we can disregard the coordinate pairs in between second and fifth, resulting in only four pairs of coordinates. Fig. 32 shows the overall visualization of red class classification.

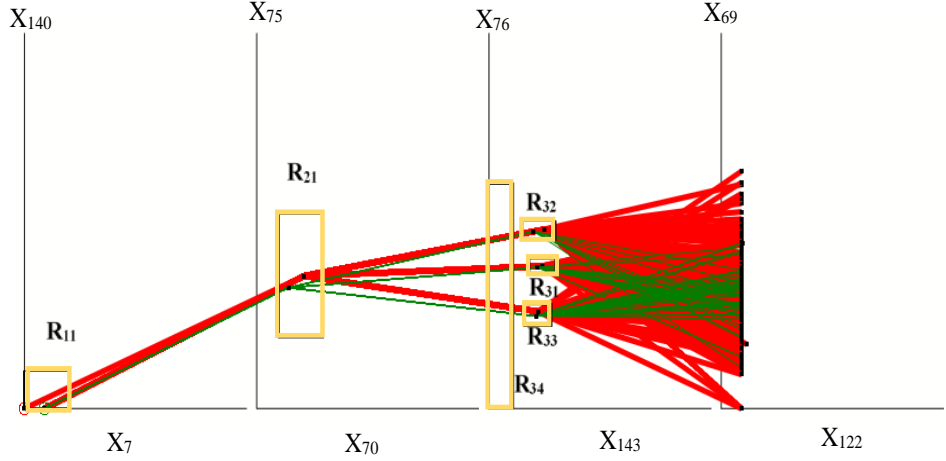

**Fig. 32.** Visualization of rule **r₁** for red class classification in the APS truck data set.

The area $R_1$ generated for red class 2 (red) classification is defined below:

$$R_1 = R_{11} \text{ \& } R_{21} \text{ \& } (R_{51} \text{ or } R_{52} \text{ or } R_{53}) \text{ \& } \neg R_{54} \tag{38}$$

The rule **r₁** for class 2 (red) classification is defined below.

$$\textbf{r}_1: \text{If } (x_7, x_{140}, x_{70}, x_{75}, x_{143}, x_{76}) \in R_1, \text{ then } \mathbf{x} \in \text{ class 2} \tag{39}$$

Data that do not follow **r₁** rule are sent to the next iteration. The Coordinate Order Optimizer is run on 166 coordinates again to get their optimized order for the remaining data. The resulting order of coordinates is $X_{61}$, $X_{164}$, $X_{103}$, $X_{97}$, $X_9$, $X_{39}$, $X_1$ and $X_{26}$. The data are visualized in Fig. 33a.

In the second iteration, the green class tends to be clustered at the bottom and red towards the top. Since the separation along vertical coordinates is clearly visible, we performed the non-linear scaling to get better data interpretation with thresholds 0.15 on $X_{164}$, 0.2 on $X_{97}$, 0.25 on $X_{39}$, and 0.25 on $X_{26}$. Then the data are visualized with non-linear scaling and analytical rules discovered (see Fig. 33b).

The area $R_2$ for class 2 (red) classification is defined below:

$$R_2 = T_1 \text{ \& } T_2 \text{ \& } T_3 \text{ \& } (T_4 \text{ or } R_{41}) \tag{40}$$

The rule **r₂** for APS data classification is given below:

$$\textbf{r}_2: \text{If } (x_{61}, x_{164}, x_{108}, x_{97}, x_9, x_{39}, x_1, x_{26}) \in R_2, \text{ then } \mathbf{x} \in \text{ class 2 else class 1.} \tag{41}$$
$$(T_1, T_2, T_3 \text{ and } T_4 \text{ are the threshold values of non-linear scaling})$$

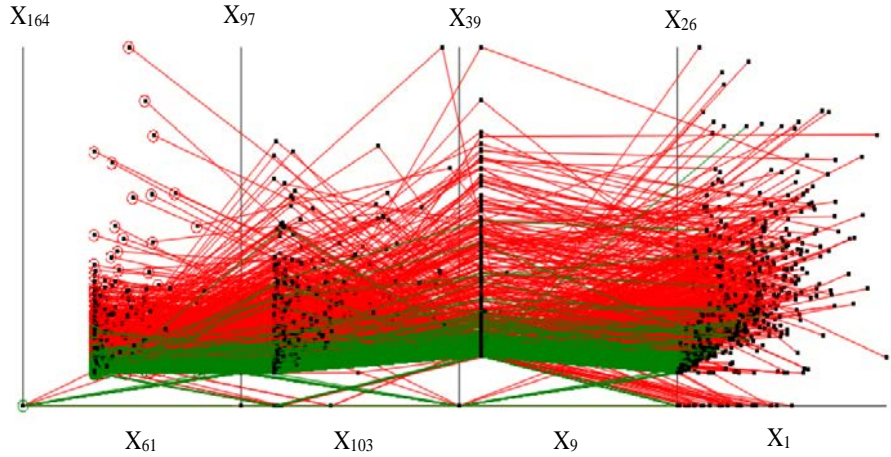

(**a**). Visualization of APS truck data set with top 8 coordinates.

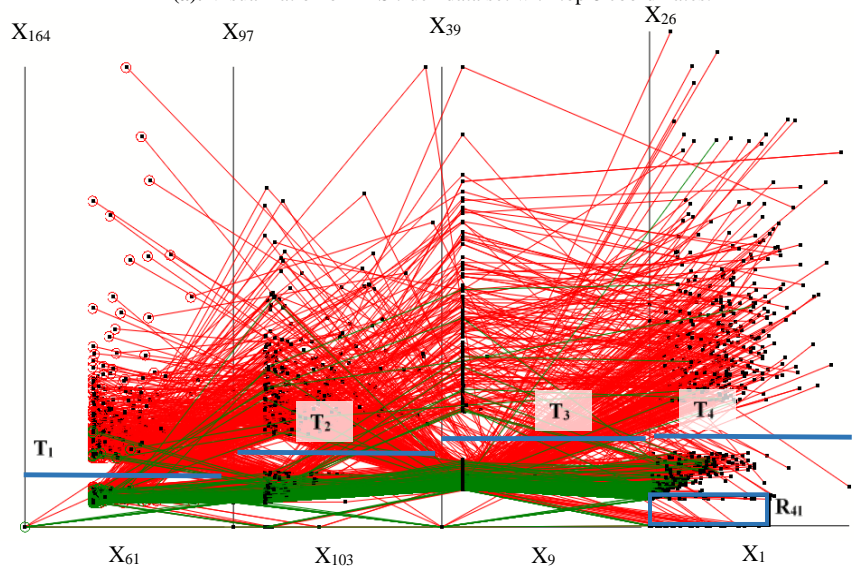

(**b**). Visualization of **r₂** in APS truck data set with top 8 coordinates with non-linear scaling.

**Fig. 33.** Visualization of APS truck data set in in the second iteration.

Table 6 lists the area parameters used in both iterations for APS data classification.

Table 6. Parameters of the rectangles  generated for classification in APS data.

| Rectangle | Left | Right | Bottom | Top | Coordinate Pair |
|---|---|---|---|---|---|
| $R_{11}$ | 0.0 | 0.2 | 0.0 | 0.1 | $(X_7, X_{140})$ |
| $R_{21}$ | 0.1 | 0.3 | 0.19 | 0.52 | $(X_{70}, X_{75})$ |
| $R_{31}$ | 0.18 | 0.3 | 0.36 | 0.4 | $(X_{143}, X_{76})$ |
| $R_{32}$ | 0.15 | 0.3 | 0.45 | 0.5 | $(X_{143}, X_{76})$ |
| $R_{33}$ | 0.16 | 0.28 | 0.22 | 0.28 | $(X_{143}, X_{76})$ |
| $R_{34}$ | 0.0 | 0.11 | 0.0 | 0.6 | $(X_{143}, X_{76})$ |
| $R_{41}$ | 0.0 | 0.6 | 0.0 | 0.1 | $(X_1, X_{26})$ |

## 7. Experimental Results and Comparison with Published Results

The results obtained are compared with the published results that uses both black-box and interpretable techniques (see Table 7).

Table 7. Comparison of Different Classification Models.

| Classification Algorithms | Accuracy % |
|---|---|
| **Breast Cancer data (9-D)** | |
| **Iterative Visual Logical Classifier (Automated)** | **99.71** |
| **Iterative Visual Logical Classifier (Interactive)** | **99.56** |
| SVM [Christobel, 2011] | 96.995 |
| DCP/RPPR [Neuhaus & Kovalchuk, 2019] | 99.3 |
| SVM/C4.5/kNN/Bayesian [Salama et al, 2012] | 97.28 |
| **Iris Data (4-D)** | |
| **Iterative Visual Logical Classifier (Automated)** | **100** |
| **Iterative Visual Logical Classifier (Interactive)** | **100** |
| Multilayer Visual Knowledge discovery [Kovalerchuk, 2018] | 100 |
| k-Means + J48 classifier [Kumar et al, 2011] | 98.67 |
| Neural Network [Swain et al, 2012] | 96.66 |
| **Seeds Data (7-D)** | |
| **Iterative Visual Logical Classifier (Automated)** | **100** |
| **Iterative Visual Logical Classifier (Interactive)** | **100** |
| Deep Neural Network [Eldem, 2020] | 100 |
| K- nearest neighbor [Sabanc et al, 2016] | 95.71 |
| **APS Failure at Scania Trucks (170-D)** | |
| **Iterative Visual Logical Classifier (Automated)** | **99.36** |
| Deep Neural Network (DNN) [Zhou et al, 2020] | 99.50 |
| Random Forest [Rafsunjani et al, 2020] | 99.025 |
| Support Vector Machine (SVM) [Rafsunjani et al, 2020] | 98.26 |

From Table 7, we can see that the classification accuracy obtained with the proposed method is on par with the published results and in some cases, have performed better than the published results. Since the interactive technique is more challenging for classifying data of larger size, we used only automated classification for such dataset (APS truck data). The results produced in this chapter are listed in bold.

From the results in Table 7, we can clearly see that the accuracies obtained from our proposed method is better than black box machine learning models [Swain et al 2012] [Eldem, 2020; Zhou et al, 2020] and on par with interpretable models [Kovalerchuk, 2018]. However, the accuracy for APS failure at Scania Trucks is slightly lesser compared to the accuracy in [Zhou et al, 2020] using Deep Neural Network, which is a black box model. Despite of lesser accuracy, our proposed model is favorable due to its transparency, the ability to use the model as self-service and the ability to interpret the model by non-technical end users.

## 8. Conclusion

In this chapter, we demonstrated the power of lossless data visualization in our proposed interpretable data classification techniques that are implemented both interactively and automatically. We observed that the interactive data classification technique works well for data with lesser cases and dimensions but fail to perform well for data with higher number of cases like the APS failure truck dataset. High degree of occlusion was observed and was challenging to discover pattern interactively. This issue was successfully addressed by our newly proposed automated interpretable technique using *Coordinate Order Optimizer* (COO) Algorithm and *Genetic Algorithm* (GA) where the areas were generated automatically rather than interactively.

We also demonstrated the power of interactive features that improved the visualization due to which discovering patterns in the data became much easier. With non-linear scaling, zooming and averaging, the visualization was improved to increase data interpretability. The SPCVis software successfully visualized the larger dataset using modified Shifted Paired Coordinates System called as Serpent Coordinate System (SCS).

Our proposed techniques can be further leveraged by incorporating more interactive features like non-orthogonal coordinates, data reversing etc. Shifted Paired Coordinates (SPC) and Serpent Coordinate System (SCS) visualization helps us to discover only specific patterns in the data and our future goal is to incorporate more General Line Coordinate Visualizations.